\documentclass[lettersize,journal]{IEEEtran}
\usepackage{amsmath,amsfonts}
\usepackage{algorithmic}
\usepackage{algorithm}
\usepackage{array}
\usepackage{textcomp}
\usepackage{stfloats}
\usepackage{url}
\usepackage{verbatim}
\usepackage{cite}
\usepackage{times}
\usepackage{soul}
\usepackage{url}
\usepackage[hidelinks]{hyperref}
\usepackage[utf8]{inputenc}
\usepackage{amsmath}
\usepackage{amsthm}
\usepackage{booktabs}
\usepackage{algorithm}
\usepackage[switch]{lineno}
\usepackage{wrapfig}
\usepackage{graphicx}
\usepackage{subcaption}
\newtheoremstyle{mytheoremstyle}
  {\topsep}       % Space above
  {\topsep}       % Space below
  {\itshape}      % Body font
  {}              % Indent amount
  {\bfseries}     % Theorem head font
  {.}             % Punctuation after theorem head
  { }             % Space after theorem head (option 2: \newline)
  {\thmname{#1}\thmnumber{ #2}\thmnote{ (#3)}} % Theorem head spec (can be left empty, meaning 'normal')
  
\newtheoremstyle{myassumptionstyle}
  {\topsep}       % Space above
  {\topsep}       % Space below
  {\itshape}              % Body font
  {}              % Indent amount
  {\bfseries}      % Assumption head font
  {.}             % Punctuation after assumption head
  { }             % Space after assumption head (option 2: \newline)
  {\thmname{#1}\thmnumber{ #2}\thmnote{ (#3)}} % Assumption head spec

\newtheoremstyle{mylemmastyle}
  {\topsep}      
  {\topsep}     
  {\itshape}      
  {}              
  {\bfseries}     
  {.}            
  { }             
  {\thmname{#1}\thmnumber{ #2}\thmnote{ (#3)}} 

\theoremstyle{mytheoremstyle}
\newtheorem{theorem}{Theorem}

\theoremstyle{myassumptionstyle}
\newtheorem{assumption}{Assumption}

\theoremstyle{mylemmastyle}
\newtheorem{lemma}{Lemma}

\begin{document}

\title{Bayesian Personalized Federated Learning with Shared and Personalized Uncertainty Representations}

\author{Hui Chen, Hengyu Liu, Longbing Cao,~\IEEEmembership{Senior Member,~IEEE,} and Tiancheng Zhang
        % <-this % stops a space
%\thanks{This paper was produced by the IEEE Publication Technology Group. They are in Piscataway, NJ.}% <-this % stops a space
\thanks{Hui Chen and Longbing Cao are with the School of Computing, Macquarie University, NSW 2109, Australia. (email: hui.chen2@students.mq.edu.au, longbing.cao@mq.edu.au)

Hengyu Liu and Tiancheng Zhang are with the School of Computer Science and Engineering, Northeastern University, 110819, China. (email: hengyuliu94@gmail.com, tczhang@mail.neu.edu.cn)}}

\markboth{Journal of \LaTeX\ Class Files,~Vol.~14, No.~8, August~2021}%
{Shell \MakeLowercase{\textit{et al.}}: A Sample Article Using IEEEtran.cls for IEEE Journals}


\maketitle

\begin{abstract}
Bayesian personalized federated learning (BPFL) addresses challenges in existing personalized FL (PFL). BPFL aims to quantify the uncertainty and heterogeneity within and across clients towards uncertainty representations by addressing the statistical heterogeneity of client data. In PFL, some recent preliminary work proposes to decompose hidden neural representations into shared and local components and demonstrates interesting results. However, most of them do not address client uncertainty and heterogeneity in FL systems, while appropriately decoupling neural representations is challenging and often ad hoc. In this paper, we make the first attempt to introduce a general BPFL framework to decompose and jointly learn shared and personalized uncertainty representations on statistically heterogeneous client data over time. A Bayesian federated neural network BPFed instantiates BPFL by jointly learning cross-client shared uncertainty and client-specific personalized uncertainty over statistically heterogeneous and randomly participating clients. We further involve continual updating of prior distribution in BPFed to speed up the convergence and avoid catastrophic forgetting. Theoretical analysis and guarantees are provided in addition to the experimental evaluation of BPFed against the diversified baselines. 
\end{abstract}

\begin{IEEEkeywords}
Federated learning, Bayesian inference, variational inference.
\end{IEEEkeywords}

\section{Introduction}
\IEEEPARstart{F}{ederated} Learning (FL) aims to solve a machine learning task in a decentralized learning manner \cite{mcmahan2017communication,DBLP:journals/tnn/SattlerKRS23,DBLP:journals/tnn/CaoZC23}. Specifically, multiple clients are coordinated by a server to collaboratively train a global model without sharing local data. FL shows advantages in privacy-preserving learning over a  distributed setting and achieves good performance when client data are identical. However, real-world clients have data showing strong statistical heterogeneity, i.e., clients are not identically distributed (non-ID) \cite{zhao2018federated,C22beyiid}. 
This requires non-IID FL and could cause potential severe performance degradation of those IID FL methods \cite{zhao2018federated,Li2020On}. 
Accordingly, personalized federated learning (PFL) \cite{fallah2020personalized,collins2021exploiting,zhang2022personalized} has recently emerged as a promising direction to address client heterogeneity by personalizing a model or its parameters for each client. 

Most PFL methods incorporate specific mechanisms or settings to address client heterogeneity under particular learning tasks. They can be roughly categorized into: multi-task learning \cite{DBLP:conf/nips/SmithCST17}, meta-learning \cite{fallah2020personalized,al2021data}, transfer learning \cite{mansour2020three}, and regularization \cite{huang2021personalized,t2020personalized} oriented PFL methods. 
Their common goal is to strike a balance between the global model and personalized local models over client updating. These methods involve manifold limitations. 
First, since both the global model and personalized models are updated over communications, it incurs multiplied computational costs, very costly especially for large-scale neural networks. 
Second, personalization by utilizing the entire model parameters may not be necessary and some softer approaches may need to be considered, e.g., model structure-based \cite{pillutla2022federated}. 
Finally, a PFL model may lack a strong credibility as it is difficult to ensure the balance across clients.

Then, a recent approach enables personalized federated learning with a shared-personalized feature/factor decoupling structure to tackle the above issues by introducing shared-personalized decomposition of model or factor structure \cite{arivazhagan2019federated,collins2021exploiting,pillutla2022federated,kotelevskii2022fedpop}. A local network is decomposed into a \textit{shared part} and a  \textit{personalized part} for each client. The parameters of the shared part are uploaded to the server for aggregation while the parameters of the personalized part remain locally updated. These methods enhance client personalization with more affordable costs. However, the existing decoupling methods of neural hidden representations are ad hoc and it is challenging to pursue an appropriate shared-personalized representation decoupling. Most of them also lack uncertainty quantification and credibility and perform poorly on small data. In addition, different shared-personalized representation decoupling methods involve distinct or ad hoc mechanisms on neural networks, e.g., FedPer \cite{arivazhagan2019federated} and FedRep \cite{collins2021exploiting}, or statistical hidden variables e.g., FedSOUL \cite{kotelevskii2022fedpop}, without a unified and general framework. 

In this paper, we make the first BPFL attempt towards robust shared-personalized uncertainty representations to address statistical heterogeneity and uncertainty of non-IID clients. In a probabilistic manner, we introduce a general BPFL framework that assumes each client can be represented by a cross-client representation with their factors shared across all clients and a client-specific representation with personalized factors for a participating client. 
Then, we instantiate this BPFL framework by a Bayesian federated neural network BPFed with both shared and personalized factors describing each client for PFL and random client participation in communications. Different from the existing Bayesian neural network (e.g., FedSOUL), the parameters of both shared and personalized models are probabilistic instead of point value-based, i.e., their parameters follow various probability distributions, and different clients own distinct personalized distributions. The distribution parameters of both shared and personalized factors are updated over communications. We further show that existing shared-personalized decomposition PFL methods are special cases of BPFed through tailoring model structure or parameter settings, i.e., BPFed represents a general PFL framework for decomposing client representations into cross-client and client-specific probabilistic components.

In summary, this work makes manifold contributions. First, a robust Bayesian PFL framework BPFed enables shared-personalized uncertainty representation learning, which quantifies client uncertainty and heterogeneity and suits small-scale data well. Then, we show BPFed serves as a general framework for shared-personalized decomposition PFL and can be customized into the existing methods. Further, we provide a continual prior updating mechanism and theoretical guarantee with a generalization bound to ensure the learning performance over iterations. Lastly, substantial experiments demonstrate the BPFed efficacy in comparison with various FL baselines, especially on small-scale data. We anticipate that the BPFed framework could benefit FL applications with small data \cite{snell2021bayesian,jospin2022hands}, novel and dynamic clients.

\section{Related Work}
\textbf{Bayesian federated learning.} To enable uncertainty quantification and more stable modeling performance, Bayesian federated learning incorporates  Bayesian learning principles into the FL framework \cite{cao2023bayesian}. BFL can be categorized into client-based and server-based methods. For client-based BFL, \cite{bhatt2022bayesian} and \cite{zhang2022personalized} introduced  BNN into FL using Markov Chain Monte Carlo and variational inference, respectively. Different from our work, these models do not decompose model structures for personalization. By applying Bayesian optimization to FL, \cite{dai2020federated} updated local models using Thompson sampling and \cite{zang2022traffic} chose expected improvement for weight adjustment. Another set of methods incorporate Bayesian nonparametric into FL settings, such as FedCor \cite{tang2022fedcor} and pFedGP \cite{achituve2021personalized} for client selection and classification based on Gaussian process, and matching algorithms \cite{yurochkin2019bayesian}, \cite{wang2020federated} for different neural network structures with Beta-Bernoulli processes. On the server side, FedBE \cite{chen2021fedbe} constructs a Gaussian or Dirichlet distribution on the server for  Bayesian model ensemble. Using Bayesian posterior decomposition, FedPA \cite{al2021federated} and QLSD \cite{vono2022qlsd} decompose the global posterior distribution into the product of local posterior distributions. However, these models assume a uniform prior and data independence across clients.

\textbf{Personalized federated learning decoupling shared from personalized parts.} Different from normal PFL methods aiming for personalization over the entire model parameters, some recent PFL networks decouple shared from personalized components of client representations. They require only part of the model parameters to gain more flexible personalized learning. FedPer \cite{arivazhagan2019federated} is an early-stage PFL with shared-personalized layer decoupling. Similar to our method, FedPer decomposes the model into common and personalized parts and update both parts simultaneously for each client. FedRep \cite{collins2021exploiting} is an alternate to FedPer with theoretical analysis. They are shown to perform similarly with both simultaneous and alternating updating mechanisms under strict convergence guarantees \cite{pillutla2022federated}. Differently, LG-FedAvg \cite{liang2019think} has an exactly opposite decoupling structure to FedPer and FedRep. Although these approaches address data heterogeneity to some extent, their performance degrades on small data and they do not handle uncertainty quantification. Recently, FedPop \cite{kotelevskii2022fedpop} decomposes the model into a fixed common part learning common representation and a random personalized part for uncertainty quantification. Compared to our method, on the one hand, the common part of FedPop lacks the ability of uncertainty quantification due to fixing common features for all clients. On the other hand, the shared prior of personalized features makes FedPop weak in learning heterogeneous architectures across clients.

\section{Problem formulation}

In this section, we provide the background of FL and PFL, and then the  formulation of BPFed for Bayesian PFL with shared-personalized uncertainty representations and statistical heterogeneity.

\textbf{Federated learning.} Assume an FL system with $N$ clients and a server for communication, the goal of FL is to solve the following optimization problem \cite{chen2022bridging}:
\begin{equation}
\min _{\mathbf{w}}\left\{F(\mathbf{w}):=\frac{1}{N} \sum_{i=1}^N F_i(\mathbf{w})\right\},
\end{equation}
\text{where} 
\begin{equation}
F_i(\mathbf{w}):=\mathbb{E}_{(\boldsymbol{x}_j,y_j)}[\mathcal{L}(\mathbf{w};\boldsymbol{x}_j,y_j)].
\end{equation}

Here, $\mathbf{w}$ denotes model parameters. $F_i(\mathbf{w})$ represents the expected risk of client $i$. $(\boldsymbol{x}_j,y_j)$ is a data instance which is drawn from the distribution of client $i$. $\mathcal{L}(\mathbf{w};\boldsymbol{x}_j,y_j)$ is the loss function for the data instance. In real scenarios, the local data of different clients usually present statistical heterogeneity due to their distinct preferences, backgrounds, and environments. It is thus challenging to train a robust model on heterogeneous client data fitting characteristics of all clients.

\textbf{Personalized federated learning.} PFL addresses client heterogeneity with the following general optimization goal: 
\begin{equation}
\begin{aligned}
\label{equation 2}
\min _{\mathbf{W}}\left\{F(\mathbf{W}):=\frac{1}{N} \sum_{i=1}^N \left(F_i(\mathbf{w}_i)+\frac{\lambda}{2} \|\mathbf{w}_i-\boldsymbol{\Omega}\|^2\right)\right\}.
\end{aligned}
\end{equation}
Here, $\mathbf{W}=\{\mathbf{w}_1, \cdots, \mathbf{w}_N \}$ is the collection of local models for all clients. $\mathbf{w}_i$ and $\boldsymbol{\Omega}$ represent the personalized model parameters of client $i$ and the parameters of the global (reference) model, respectively. $\lambda$ is the regularization coefficient for adjusting the degree of personalization. \eqref{equation 2} implies that the goal of local updating over clients aims for a balance between the global model and the personalized client models. Consequently, PFL outperforms standard FL with better generalization and performance on heterogeneous data. However, PFL consumes more computational power and is structurally redundant because of personalization via entire model parameters. More importantly, there are no indicators measuring whether local updating has achieved the desired balance.

\begin{figure*}[t]
    \centering
    \includegraphics[width=0.8\textwidth]{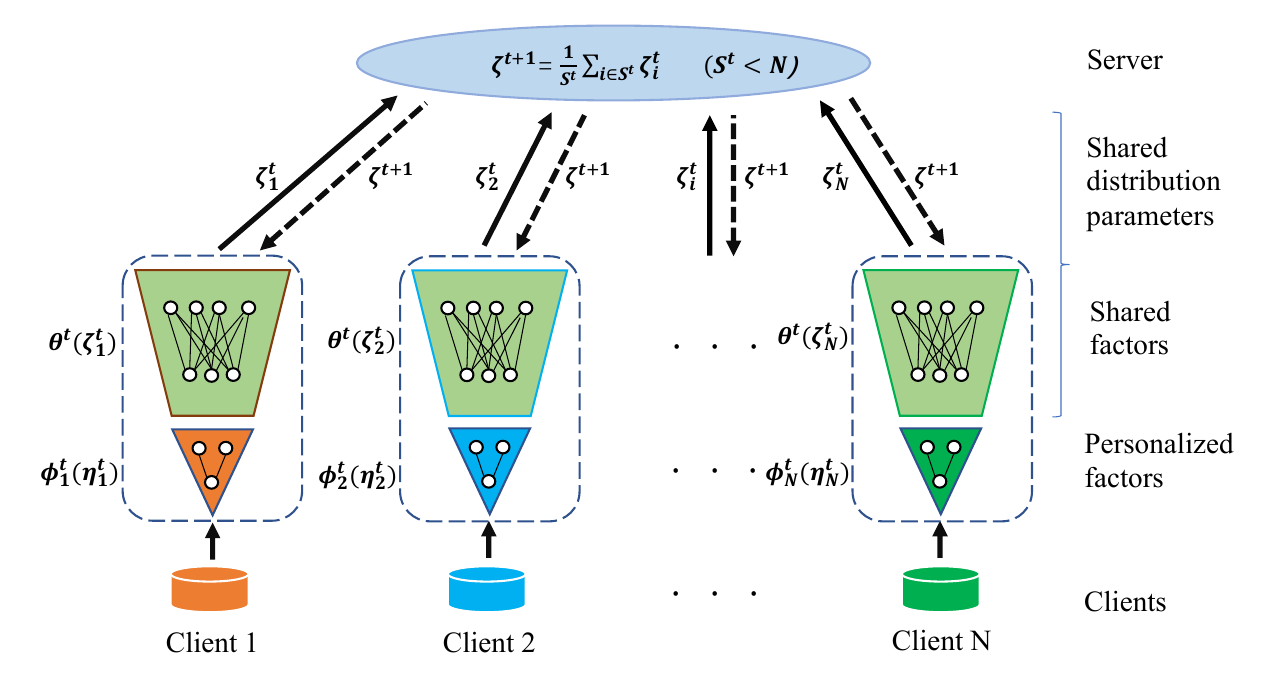}
    \caption{Architectural and statistical structure of BPFed. For client $i$, its parameters of a Bayesian federated neural network are decomposed into cross-client shared factors $\boldsymbol{\theta}$ and client-specific personalized factors $\boldsymbol{\phi}_i$. All clients upload their distribution parameters $\boldsymbol{\zeta}$ of shared factors to the server for aggregation, while the distribution parameters $\boldsymbol{\eta}_i$ of personalized factors are updated locally. We implement the shared and personalized factors by probability distributions with respective distribution parameters.}
    \label{figure 1}
\end{figure*}

\textbf{BPFed: Bayesian personalized federated learning with shared-personalized uncertainty representations.} Taking the assumption in \cite{collobert2011natural} that a multi-task learning model is decomposed of common features across edge devices in a heterogeneous data environment, we decompose a PFL's client representation into two parts: a shared  representation for all clients and a personalized representation for each client. This shared-personalized PFL framework can be instantiated into neural, probabilistic or Bayesian neural models. In the neural setting, the shared-personalized hidden feature decomposition results in a cross-client common model (or layer) and their model parameters and a client-specific personalized model (or layer) and their personalized parameters. In the statistical principle, the shared-personalized hidden factor decomposition corresponds to shared and personalized hidden factors (variables) respectively and jointly characterizing each client. As this paper focuses on BPFL, Fig. \ref{figure 1} illustrates the architectural and statistical structure of our proposed Bayesian PFL BPFed with shared and personalized factors represented by probability distributions and their distribution parameters over Bayesian neural networks.

Given a set of $N$ clients with data $\boldsymbol{D}=\sqcup_{i=1}^N \boldsymbol{D}_i$,
and $\boldsymbol{D}_i=(\boldsymbol{d}_i^1, \cdots, \boldsymbol{d}_i^n)$ denotes the data for the $i$-th client and $\boldsymbol{d}_i^j=(\boldsymbol{x}_i^j,y_i^j)$ is  the $j$-th data sample for the $i$-th client. The local marginal likelihood function $p(\boldsymbol{D}_i \mid \boldsymbol{\eta}_i, \boldsymbol{\zeta})$ for each client can be defined by
\begin{equation}
p(\boldsymbol{D}_i \mid \boldsymbol{\eta}_i, \boldsymbol{\zeta}) = \int_{\mathbb{R}^{d_{\boldsymbol{\phi}}}} \int_{\mathbb{R}^{d_{\boldsymbol{\theta}}}} p(\boldsymbol{D}_i \mid \boldsymbol{\phi}_i, \boldsymbol{\theta}) p(\boldsymbol{\phi}_i, \boldsymbol{\theta} \mid \boldsymbol{\eta}_i, \boldsymbol{\zeta}) d \boldsymbol{\phi}_i d \boldsymbol{\theta},
\end{equation}
where $\boldsymbol{\phi}_i \in \mathbb{R}^{d_{\boldsymbol{\phi}}}$ stands for the personalized factors for the $i$-th client. $\boldsymbol{\theta} \in \mathbb{R}^{d_{\boldsymbol{\theta}}}$ represents the shared factors  by all clients. 

\begin{figure}[!t]
\centering
\includegraphics[width=0.45\textwidth]{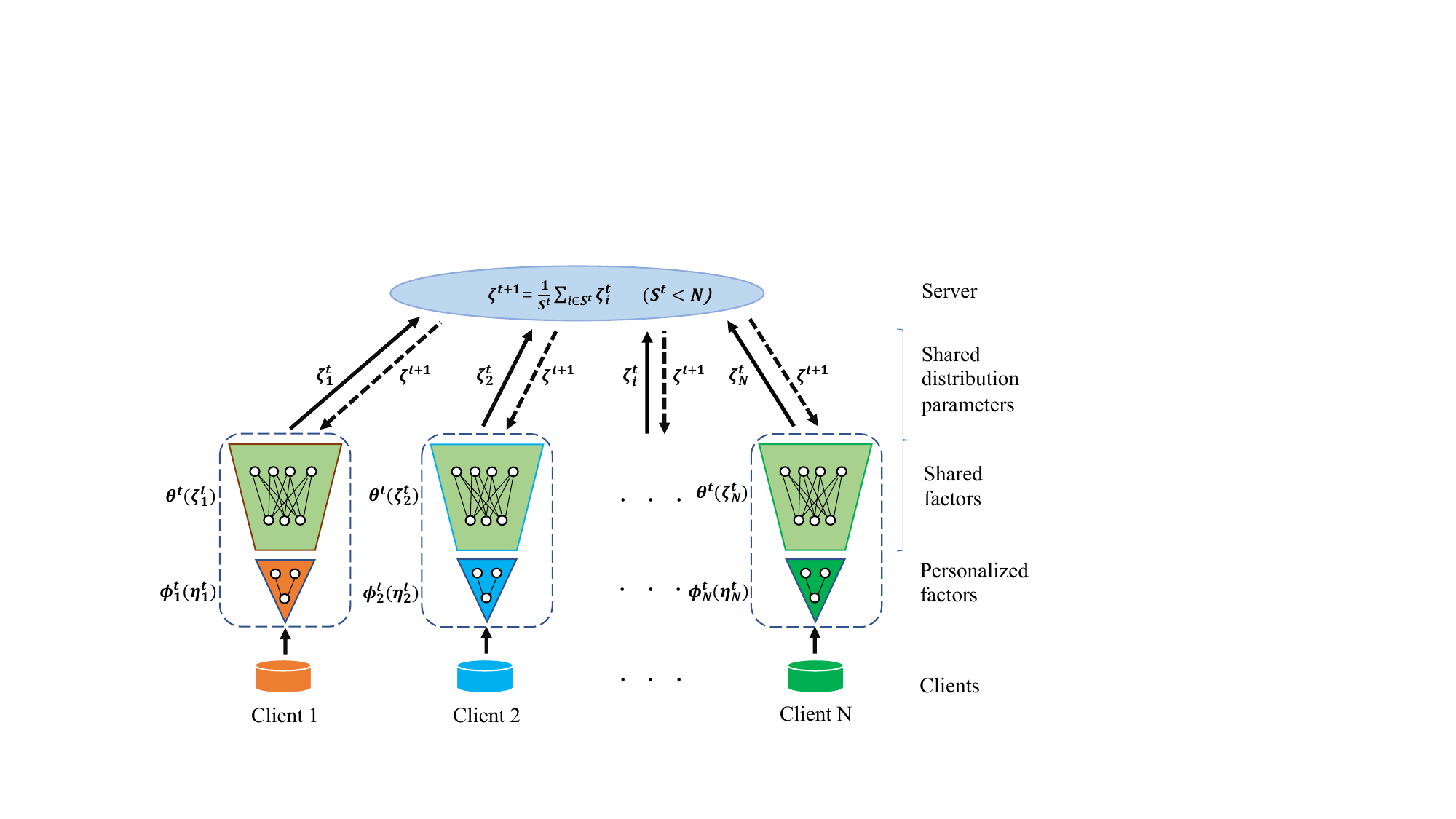}
\caption{BPFed statistical structure of cross-client shared factors and client-specific personalized factors describing clients with continual prior distribution updating.}
\label{figure 2}
\vspace{-4mm}
\end{figure}

In BPFed, $\boldsymbol{\phi}_i$ and $\boldsymbol{\theta}$ are random variables instead of specific values as in other PFL models with shared-personalized decomposition. Accordingly, $\boldsymbol{\eta_i}$ and $\boldsymbol{\zeta}$ stand for the client-specific and cross-client distribution parameters for these random variables, respectively. Fig. \ref{figure 2} shows the BPFed statistical structure of the shared and personalized factors and their dependent distribution parameters.

Here, we instantiate BPFed in a simplest way by applying the same neural network architecture to each client but differing their personalized model parameters for each client: $(\boldsymbol{\phi}_i, \boldsymbol{\theta}) \in \mathbb{R}^{d_{\boldsymbol{\phi}}+d_{\boldsymbol{\theta}}}$ ($d_{\boldsymbol{\phi}} \ll d_{\boldsymbol{\theta}}$). Furthermore, we assume the prior distribution of the $i$-th client to be  $p(\boldsymbol{\phi}_i, \boldsymbol{\theta})$. Then, we obtain the posterior distribution of model parameters for each client using the Bayes' theorem as follows:
\begin{equation}
p(\boldsymbol{\phi}_i, \boldsymbol{\theta} \mid \boldsymbol{D}_i) \propto {p(\boldsymbol{D}_i \mid \boldsymbol{\phi}_i, \boldsymbol{\theta})p(\boldsymbol{\phi}_i, \boldsymbol{\theta})}.
\end{equation}

Since it is intractable to solve the posterior distribution $p(\boldsymbol{\phi}_i, \boldsymbol{\theta} \mid \boldsymbol{D}_i)$, we approximate it utilizing variational inference \cite{blei2017variational,DBLP:journals/tnn/LanLHYC23,DBLP:journals/tnn/YeB23}, which finds the optimal variational distribution to the posterior distribution. Consequently, the objective function of BPFed is:
\begin{equation}
\label{eq.5}
\min _{\{\boldsymbol{\phi}_1, \cdots,\boldsymbol{\phi}_N,\boldsymbol{\theta}\}}\left\{F(\boldsymbol{\phi}_1, \cdots, \boldsymbol{\phi}_N,\boldsymbol{\theta}):=\frac{1}{N} \sum_{i=1}^N F_i(\boldsymbol{\phi}_i, \boldsymbol{\theta})\right\},
\end{equation}
where
\begin{equation}
\begin{aligned}
\label{eq.6}
F_i(\boldsymbol{\phi}_i, \boldsymbol{\theta}):=&\min _{q(\boldsymbol{\phi}_i,\boldsymbol{\theta})}\left\{-\mathbb{E}_{q(\boldsymbol{\phi}_i, \boldsymbol{\theta})}[\log p(\boldsymbol{D}_i\mid\boldsymbol{\phi}_i,\boldsymbol{\theta})]\right.\\
&+\left.\mathrm{KL}[q(\boldsymbol{\phi}_i, \boldsymbol{\theta}) \| p(\boldsymbol{\phi}_i, \boldsymbol{\theta})]\right\}.
\end{aligned}
\end{equation}
 Here, \eqref{eq.6} corresponds to the negative Evidence Lower Bound (ELBO). $p(\boldsymbol{D_i\mid}\boldsymbol{\phi}_i,\boldsymbol{\theta})$ and $q(\boldsymbol{\phi}_i, \boldsymbol{\theta})$ denote the likelihood function and the variational distribution associated with the model parameters for the $i$-th client, respectively. $p(\boldsymbol{\phi}_i, \boldsymbol{\theta})$ represents the prior distribution of model parameters for the $i$-th client, which will be further discussed in Section \ref{algorithm}.

\section{Algorithm}
\label{algorithm}

In this section, we will solve the optimization problems in \eqref{eq.5} and \eqref{eq.6} for BPFed through the mini-batch gradient descent optimization. We will further discuss how BPFed represents a general framework for the related PFL methods.

\textbf{Algorithm implementation.} In BPFed, in each communication round $t$, each client uploads its distribution parameters of shared factors to the server for updating, while the distribution parameters of personalized factors are only updated locally. 

Like most of related work \cite{al2021federated,chen2021fedbe,zhang2022personalized}, we assume that the model parameters follow Gaussian distribution and the variational distribution $q(\boldsymbol{\phi}_i, \boldsymbol{\theta})$ comes from a mean-field family, which suggests each model parameter is independent of the other ones. To this end, we denote $\boldsymbol{\eta}_i=(\boldsymbol{\mu}_{\boldsymbol{\eta}_i},\boldsymbol{\sigma}_{\boldsymbol{\eta}_i})$ and $\boldsymbol{\zeta}=(\boldsymbol{\mu}_{\boldsymbol{\zeta}},\boldsymbol{\sigma}_{\boldsymbol{\zeta}})$, where $\boldsymbol{\mu}_{\boldsymbol{\eta}_i}, \boldsymbol{\sigma}_{\boldsymbol{\eta}_i}$ and $\boldsymbol{\mu}_{\boldsymbol{\zeta}}, \boldsymbol{\sigma}_{\boldsymbol{\zeta}}$ represent the mean  and standard deviation vectors of the personalized  and shared factors respectively, as shown in Fig. \ref{figure 2}. 

For the first term in \eqref{eq.6}, we resort to the reparameterization trick \cite{kingma2013auto} to calculate the gradient. By introducing the random variable $\boldsymbol{\epsilon} \sim \mathcal{N}(\boldsymbol{0}, \boldsymbol{I})$, the model parameters can be obtained as follows, as shown in Fig. \ref{figure 2}:
\begin{equation}
\label{eq.7}
\boldsymbol{\phi}_i=\boldsymbol{\mu}_{\boldsymbol{\eta}_i}+\boldsymbol{\sigma}_{\boldsymbol{\eta}_i} \odot \boldsymbol{\epsilon},
\end{equation}
\begin{equation}
\label{eq.8}
\boldsymbol{\theta}=\boldsymbol{\mu}_{\boldsymbol{\zeta}}+\boldsymbol{\sigma}_{\boldsymbol{\zeta}} \odot \boldsymbol{\epsilon}.
\end{equation}
Then we can obtain the expectation through Monte Carlo estimation.

For the second term in \eqref{eq.6}, we can obtain the closed-form solution of KL divergence under the mean-field assumption. Furthermore, in order to avoid catastrophic forgetting and achieve a fast convergence rate, we adopt continual learning to construct the prior distribution $p(\boldsymbol{\phi}_i, \boldsymbol{\theta})$ in our setting. In particular, the personalized factors of prior distribution for the $i$-th client $p(\boldsymbol{\phi}_i^t, \cdot)$ in the current communication round $t$ inherits the updated personalized factors of the previous communication round $t-1$: $p(\boldsymbol{\phi}_i^t, \cdot) \to \pi(\boldsymbol{\phi}_i^{t-1}, \cdot)$. Then, we replace the shared factors of prior distribution for the $i$-th client $p(\cdot, \boldsymbol{\theta}^t)$ at $t$ with the aggregated result: $p(\cdot, \boldsymbol{\theta}^t) \to \pi(\cdot, \boldsymbol{\theta}^{t})$.

Consequently, at the $t$-th communication round, we represent the prior distribution as $\pi(\boldsymbol{\phi}_i^{t-1}, \boldsymbol{\theta}^{t})$, and the objective function of the $i$-th client in \eqref{eq.5} can then be approximated as follows:
\begin{equation}
\begin{aligned}
\label{eq.9}
Q_i(\boldsymbol{\gamma}) \approx &-\frac{n}{b}\frac{1}{M}\sum^b_{j=1}\sum^M_{m=1}\log p_{\boldsymbol{\gamma}_m}(\boldsymbol{D}_i^j\mid\boldsymbol{\phi}_i^t,\boldsymbol{\theta}_i^{t+1})\\
&+\mathrm{KL}[q_{\boldsymbol{\gamma}}(\boldsymbol{\phi}_i^t, \boldsymbol{\theta}_i^{t+1}) \| \pi_{\boldsymbol{\gamma}}(\boldsymbol{\phi}_i^{t-1}, \boldsymbol{\theta}^{t})],
\end{aligned}
\end{equation}
where we define $\boldsymbol{\gamma}=(\boldsymbol{\eta}_i, \boldsymbol{\zeta}_i)$ for the sake of better expression. $n$ is the sample size of the $i$-th client. $b$ and $M$ represent the mini-batch size and Monte Carlo sample size, respectively. Similar to \cite{zhang2022personalized}, %considering the communication overhead, 
to reinforce the KL-divergence, we further update the pre-aggregation local model for the $i$-th client in each iteration by 

\begin{equation}
\label{eq.10}
\overline{Q_i(\boldsymbol{\gamma})}= \mathrm{KL}[q_{\boldsymbol{\gamma}}(\boldsymbol{\phi}_i^t, \boldsymbol{\theta}_i^{t+1}) \| \pi_{\boldsymbol{\gamma}}(\boldsymbol{\phi}_i^{t-1}, \boldsymbol{\theta}^{t})].
\end{equation}

To update the model parameters of shared factors and personalized factors, different strategies may be taken, e.g., simultaneously updating both parts  or alternately updating each, for each client. Given that the latter method requires more upload bandwidth and there may not be significant performance difference between these two updating approaches \cite{pillutla2022federated,kotelevskii2022fedpop,DBLP:journals/tnn/ShahL23}, we  simultaneously update the shared and personalized factors for BPFed.

Further, we apply BPFed to a more realistic dynamic cross-device federated learning setting, where a random subset $S^t$ with size $S$ ($S \le N$) of clients may participate in each communication round $t ~( \le T)$. This captures the randomness and dynamics of client participation in FL tasks. Accordingly, the server averages the updated distribution parameters of shared factors for those participating clients to obtain the new shared factors. 
Algorithm \ref{alg.1} illustrates the BPFed training process. 

\begin{algorithm}
	%\textsl{}\setstretch{1.8}
	\renewcommand{\algorithmicrequire}{\textbf{Server executes:}}
	\renewcommand{\algorithmicensure}{\textbf{Client} Update($i, \boldsymbol{\eta}_i^{t-1}, \boldsymbol{\zeta}^{t}$)$\boldsymbol{:}$}
	\caption{\texttt{BPFed}: Bayesian personalized federated learning with shared-personalized factor decomposition and their joint uncertainty representations}
	\label{alg.1}
	\begin{algorithmic}%[1]
	\STATE \textbf{Input:} $T$ communication rounds, $R$ local updating rounds, $S$ random subsets of participating clients, and initialization: $\boldsymbol{\eta}_i^0=(\boldsymbol{\mu}_{\boldsymbol{\eta}_i}^0,\boldsymbol{\sigma}_{\boldsymbol{\eta}_i}^0)$, $\boldsymbol{\zeta}^0=(\boldsymbol{\mu}_{\boldsymbol{\zeta}}^0,\boldsymbol{\sigma}_{\boldsymbol{\zeta}}^0)$
       \REQUIRE
	\FOR{$t=0,1,\dots, T-1$}
        \STATE Sample $\mathcal{S}^t$ clients with size $S$ uniformly at random
        \FOR{each client $i \in \mathcal{S}^t$ \textbf{in parallel}}
        \STATE $\boldsymbol{\eta}_i^{t}, \boldsymbol{\zeta}_i^{t+1} \gets$ Client Update($i, \boldsymbol{\eta}_i^{t-1}, \boldsymbol{\zeta}^{t}$)
        \ENDFOR
        \STATE$\boldsymbol{\zeta}^{t+1}=\frac{1}{{S}}\sum_{i \in \mathcal{S}^t}\boldsymbol{\zeta}_i^{t+1}$
        \ENDFOR
        \ENSURE
        \STATE $\boldsymbol{\eta}_{i,0}^{t} \gets \boldsymbol{\eta}_i^{t-1}$, $\boldsymbol{\zeta}_{i,0}^{t} \gets \boldsymbol{\zeta}^{t}$
        \FOR{$r=0,1, \dots, R-1$}
        \STATE Sample a mini-batch with size b from $\boldsymbol{D}_i$
        \STATE Draw samples from $\boldsymbol{\epsilon} \sim \mathcal{N}(\boldsymbol{0}, \boldsymbol{I})$
        \STATE Update $\boldsymbol{\eta}$, $\boldsymbol{\zeta}$ with $\nabla Q_i(\boldsymbol{\gamma})$ in Eq. (\ref{eq.9}) using mini-batch gradient descent
          \STATE Update $\overline{\boldsymbol{\eta}}$, $\overline{\boldsymbol{\zeta}}$ with $\nabla \overline{Q_i(\boldsymbol{\gamma})}$ in Eq. (\ref{eq.10}) using mini-batch gradient descent
        \ENDFOR 
        
        \textbf{Return} $\overline{\boldsymbol{\eta}_{i,R}^{t}}$ $\overline{\boldsymbol{\zeta}_{i,R}^{t+1}}$
	\end{algorithmic}
\end{algorithm}

\textbf{Generalizing BPFed for existing PFL methods.} By customizing the parameters and/or model structure settings, BPFed can be customized into the existing PFL methods with shared-personalized decomposition. Specifically, when $\boldsymbol{\sigma}_{\boldsymbol{\eta}_i} \to \boldsymbol{0}^+$ and $\boldsymbol{\sigma}_{\boldsymbol{\zeta}} \to \boldsymbol{0}^+$, where $\boldsymbol{\eta}_i$ and $\boldsymbol{\zeta}$ can be regarded as two Dirac distributions centered at $\boldsymbol{\mu}_{\boldsymbol{\eta}_i}$ and $\boldsymbol{\mu}_{\boldsymbol{\zeta}}$, which also means that the model loses the ability of uncertainty quantification. In this way, BPFed can be customized to FedPer (updating $\boldsymbol{\mu}_{\boldsymbol{\eta}_i}$ and $\boldsymbol{\mu}_{\boldsymbol{\zeta}}$ simultaneously) \cite{arivazhagan2019federated}, FedRep (updating $\boldsymbol{\mu}_{\boldsymbol{\eta}_i}$ and $\boldsymbol{\mu}_{\boldsymbol{\zeta}}$ alternately) \cite{collins2021exploiting}, and LG-FedAvg (by model structure transformation) \cite{liang2019think}. In addition, when $\boldsymbol{\sigma}_{\boldsymbol{\zeta}} \to \boldsymbol{0}^+$ and $\boldsymbol{\eta}_i$ is the same for all clients, which suggests that the statistical heterogeneity of data among the clients can be only reflected by $\boldsymbol{\sigma}_{\boldsymbol{\eta}_i}$. In this way, BPFed is customized into FedPop \cite{kotelevskii2022fedpop}. Therefore, BPFed can be considered as a unified general framework of PFL with shared-personalized factor/model decomposition.

\section{Theoretical analysis}

In this section, we present a detailed theoretical analysis of our method. An upper bound of average generalization error is given under the Hellinger distance for all clients. % and show the rate-minimax convergence up to a logarithmic term.
The proofs of main lemmas are given in the Appendix.

Assume a neural network with $L$ hidden layers is given for each client, the number of neurons on the $l$-th hidden layer is $e_l$ ($l=1,\dots,L$), thus the network has a neuron set $\boldsymbol{e}=(e_1, \cdots, e_L)$. Besides, assume $f_i$ and $f_{\boldsymbol{\phi}_i, \boldsymbol{\theta}}$ denote the true function and predicted function of each data instance for the $i$-th client, respectively, their corresponding density functions are represented by $p_i$ and $p_{\boldsymbol{\phi}_i, \boldsymbol{\theta}}$. We further assume there are $T_1$ personalized factors $\boldsymbol{\phi_i}$ and $T_2$ shared factors $
\boldsymbol{\theta}$ associated with client $i$. With the above settings, the client networks are of equal width \cite{polson2018posterior,bai2020efficient} and follow the following assumptions.
\begin{assumption}
Equal-width Bayesian neural network with width $e_l=K$.
\end{assumption}
\begin{assumption}
The activation function $g(\cdot)$ of Bayesian neural network is 1-Lipschitz continuous.
\end{assumption}
\begin{assumption}
Each model parameter is upper bounded by $B$ (i.e. $\|\boldsymbol{\phi}_i, \boldsymbol{\theta}\|_{\infty} \leq B$).
\end{assumption}
%\textbf{H1.} \textit{Equal-width Bayesian neural network with width $e_l=K$.}
%\textbf{H2.} \textit{The activation function $g(\cdot)$ of Bayesian neural network is 1-Lipschitz continuous.}
%\textbf{H3.} \textit{Each model parameter is upper bounded by $B$ (i.e. $\|\boldsymbol{\phi}_i, \boldsymbol{\theta}\|_{\infty} \leq B$).}

%It should be noted that while the above assumptions simplify our model, they also render our research problem amenable to analytical solutions. 

Based on above assumptions, we rewrite the optimization problem in \eqref{eq.6} for each client based on the above assumptions as follows:
\begin{equation}
\begin{aligned}
F_i(\boldsymbol{\phi}_i, \boldsymbol{\theta}):=&\min _{q(\boldsymbol{\phi}_i,\boldsymbol{\theta})}\left\{\int_{\Theta}\int_{\Theta} \ell_n(P_i,P_{\boldsymbol{\phi}_i, \boldsymbol{\theta}})q(\boldsymbol{\phi}_i,\boldsymbol{\theta})d \boldsymbol{\phi}_i d \boldsymbol{\theta}\right.\\
&+\left.\mathrm{KL}[q(\boldsymbol{\phi}_i, \boldsymbol{\theta}) \| \pi(\boldsymbol{\phi}_i, \boldsymbol{\theta})]\right\},
\end{aligned}
\end{equation}
where $\ell_n(P_i,P_{\boldsymbol{\phi}_i, \boldsymbol{\theta}})=\log\frac{p_i(\boldsymbol{D}_i)}{p_{\boldsymbol{\phi}_i,\boldsymbol{\theta}}(\boldsymbol{D}_i)}$, denoting the log-likelihood ratio between $p_i$ and $p_{\boldsymbol{\phi}_i,\boldsymbol{\theta}}$. $P_i$ and $P_{\boldsymbol{\phi}_i, \boldsymbol{\theta}}$ represent the underlying probability measure of data.

We give the following error measures:
\begin{align}
&r^n:=((L+1) (T_1+T_2) / n) \log N \nonumber \\
&\ \ \ \ \ \ \ +((T_1+T_2) / n) \log (e_0 \sqrt{n / (T_1+T_2)}),\\
&\xi_i^n:=\inf _{\boldsymbol{\phi}_i, \boldsymbol{\theta} \in \Theta\left(L, \boldsymbol{e}\right),\|\boldsymbol{\phi}_i, \boldsymbol{\theta}\|_{\infty} \leq B}\left\|f_{\boldsymbol{\phi}_i, \boldsymbol{\theta}}-f_i\right\|_{\infty}^2,\\
& \varepsilon^n:=\alpha n^{-1 / 2} \log ^\delta(n) \nonumber\\
& \sqrt{(T_1+T_2) \bigg ((L+1) \log K+\log \left(e_0 \sqrt{\frac{n}{T_1+T_2}}\right)\bigg )},
\end{align}
where $r^n$ is the estimation error of variational inference approximation and $\varepsilon^n$ is the estimation error from the statistacal estimator of $P_i$, while $\xi_i^n$ is the approximation error between $f_{\boldsymbol{\phi}_i, \boldsymbol{\theta}}$ and $f_i$. The input dimensionality is denoted by $e_0$. $\alpha$ is a large constant and $\delta>1$. 

\begin{lemma}
Let both Assumptions 1 and 2 hold, then we have the following inequality with a dominating probability,
\begin{equation}
\begin{aligned}
\frac{1}{N}\sum_{i=1}^{N}\inf _{q(\boldsymbol{\phi}_i,\boldsymbol{\theta})}\bigg\{\int_{\Theta}\int_{\Theta} \ell_n(P_i,P_{\boldsymbol{\phi}_i, \boldsymbol{\theta}})q(\boldsymbol{\phi}_i,\boldsymbol{\theta})d \boldsymbol{\phi}_i d \boldsymbol{\theta}\\
+\mathrm{KL}[q(\boldsymbol{\phi}_i, \boldsymbol{\theta}) \| \pi^*(\boldsymbol{\phi}_i, \boldsymbol{\theta})]\bigg\}
\leq nC r^n +\frac{nC'}{N}\sum_{i=1}^N \xi_i^n.
\end{aligned}
\end{equation}
Here, $\pi^*(\boldsymbol{\phi}_i, \boldsymbol{\theta})$ is the optimal prior distribution for the $i$-th client. $C$ and $C'$ are positive constants when $\lim n(r^n+\xi_i^n)=\infty$ or any divergent sequences when $\lim \sup n(r^n+\xi_i^n)\neq \infty$.
\end{lemma}

To investigate the average generalization error of BPFed, we first define the Hellinger distance between $P_i$ and $P_{\boldsymbol{\phi}_i,\boldsymbol{\theta}}$:
\begin{equation}
\begin{aligned}
&d^2\left(P_{\boldsymbol{\phi}_i, \boldsymbol{\theta}}, P_i\right)\\
&=\mathbb{E}_{\boldsymbol{x}}\left(1-\exp \left\{-\left[f_{\boldsymbol{\phi}_i, \boldsymbol{\theta}}(\boldsymbol{x})-f_i(\boldsymbol{x})\right]^2 /\left(8 \sigma_\epsilon^2\right)\right\}\right),
\end{aligned}
\end{equation}
where $\sigma_\epsilon^2$ denotes the variance of model noise.

Then we have Lemma 2 for the Hellinger distance under the above assumptions.

\begin{lemma}
Let Assumptions 1-3 hold, then we have the following inequality with a dominating probability,
\begin{equation}
\begin{aligned}
&\frac{1}{N}\sum_{i=1}^{N}\int_{\Theta}\int_{\Theta} d^2(P_i,P_{\boldsymbol{\phi}_i, \boldsymbol{\theta}})\hat{q}(\boldsymbol{\phi}_i,\boldsymbol{\theta})d \boldsymbol{\phi}_i d \boldsymbol{\theta} \\
\leq &\frac{1}{nN}\sum_{i=1}^{N}\inf _{q(\boldsymbol{\phi}_i,\boldsymbol{\theta})}\bigg\{\int_{\Theta}\int_{\Theta} \ell_n(P_i,P_{\boldsymbol{\phi}_i, \boldsymbol{\theta}})q(\boldsymbol{\phi}_i,\boldsymbol{\theta})d \boldsymbol{\phi}_i d \boldsymbol{\theta}
\\
&+\mathrm{KL}[q(\boldsymbol{\phi}_i, \boldsymbol{\theta}) \| \pi^*(\boldsymbol{\phi}_i, \boldsymbol{\theta})]\bigg\}+C'' (\varepsilon^n)^2,
\end{aligned}
\end{equation}
where $C''$ is a constant.
\end{lemma}

We can obtain the following theoretical guarantee for the average generalization error of all clients by combining Lemma 1 and Lemma 2.
\begin{table*}[t]
\renewcommand{\arraystretch}{1.3}
\caption{Comparison of accuracy (\%) on MNIST, FMNIST and CIFAR-10 w.r.t. small and large data size. 
  % Best results are boldfaced.
  }
\label{table 1}
\centering
\begin{tabular}{ccccccc}
\toprule
& \multicolumn{2}{c}{MNIST} & \multicolumn{2}{c}{FMNIST} & \multicolumn{2}{c}{CIFAR-10} \\
 \midrule 
Dataset size                              & Small & Large & Small & Large & Small & Large \\
 \midrule
 
 FedAvg  & $88.27\pm0.09$ & $89.32\pm0.08$ & $81.22\pm0.02$ & $82.13\pm0.67$ & $39.83 \pm0.07$& $55.23\pm0.13$ \\ 
 Per-FedAvg & $92.13\pm0.36$ & $97.67\pm1.87$ & $87.38\pm0.48$ & $92.55\pm0.56$ & $52.15\pm0.56$ & $69.68\pm0.52$ \\
 pFedMe  & $93.70\pm0.32$ & $96.44\pm0.19$ & $89.24\pm0.52$ & $91.88\pm0.58$ & $52.58\pm0.43$ & $66.80\pm0.33$ \\
 pFedBayes  & $93.31\pm0.04$ & $98.15\pm0.23$ & $87.91\pm0.12$ & $92.38\pm0.21$ & $50.39\pm0.22$ & $68.23\pm0.42$  \\
 
\midrule 
 FedPer  & $93.77\pm0.06$ & $98.49\pm0.01$ & $89.06\pm0.02$ & $93.43\pm0.16$ & $51.65\pm0.08$ & $\boldsymbol{73.99 \pm0.05}$ \\
 LG-FedAvg  & $91.59\pm0.16$ & $97.18\pm0.04$ & $86.79\pm0.16$ & $91.85\pm0.24$ & $41.87\pm0.24$ & $64.88\pm0.42$ \\
 FedRep   & $94.61\pm0.18$ & $98.50\pm 0.05$ & $88.76\pm 0.57$ & $93.15\pm0.06$ & $46.80\pm0.15$  & $72.18\pm0.03$ \\
 FedSOUL  & $91.71\pm0.10$ & $97.69\pm0.01$ & $86.63\pm0.06$ & $91.87\pm0.04$ & $43.83\pm0.23$ & $62.70\pm0.14$ \\
\midrule 
BPFed (Ours)                              & $\boldsymbol{94.75 \pm 0.34}$& $\boldsymbol{98.51 \pm0.51}$& $\boldsymbol{89.68\pm0.62}$ & $\boldsymbol{93.54\pm0.43}$ & $\boldsymbol{54.37\pm0.32}$ & $70.14\pm0.46$ \\
\bottomrule
\end{tabular}
\end{table*} 

\begin{theorem}
Assume Assumptions 1-3 hold, then we have the following upper bound of average generalization error with a dominating probability,
\begin{equation}
\begin{aligned}
\frac{1}{N}\sum_{i=1}^{N}\int_{\Theta}\int_{\Theta} d^2(P_i,P_{\boldsymbol{\phi}_i, \boldsymbol{\theta}})\hat{q}(\boldsymbol{\phi}_i,\boldsymbol{\theta})d \boldsymbol{\phi}_i d \boldsymbol{\theta}\\
\leq C r^n +\frac{C'}{N}\sum_{i=1}^N \xi_i^n+C'' (\varepsilon^n)^2,
\end{aligned}
\end{equation}
where $C$ and $C'$ are any divergent sequences; $C''$ is a positive constant.
\end{theorem}

\textbf{Remark.} \textit{Theorem 1 shows that the upper bound of average generalization error contains three error terms $r^n$, $\xi_i^n$ and $\varepsilon^n$. The first two error terms are estimation errors and the last error term is the approximation error. As we know, both types of errors only depend on the model hypothesis set (i.e., the total $T_1+T_2$ model parameters) in the case of a fixed number of samples. Recall the Assumption 1, we find that the setting of parameter numbers $T_1$ and $T_2$ (for personalized and shared factors) corresponds to different neural network structures, thus differing the local representation ability of the personalized factors and the global representation ability of the shared factors for the model. Therefore, how to set the appropriate values for $T_1$ and $T_2$ according to practical tasks is a problem for future work.}

\textbf{Remark.} \textit{The form of Theorem 1 is similar to that in the existing literature \cite{zhang2022personalized}, but our approach can provide more stable and superior convergence guarantees. Firstly, the proof of Lemma 1 relies on a optimal prior distribution. Since our client-specific prior distribution inherits the local updates' results from previous communication round, we can get closer to the optimal prior distribution, resulting in superior convergence guarantees. Additionally, in literature \cite{zhang2022personalized}, model personalization is controlled by the hyperparameter $\zeta$, and the magnitude of this hyperparameter directly affects the final convergence results. In contrast, in our approach, model personalization is obtained directly from local updates, allowing us to achieve more stable convergence guarantees.} 

\section{Experiments}

In this section, we evaluate the performance of BPFed against various types of FL methods in a heterogeneous setting.

\textbf{Datasets.} We evaluate our method against baseline methods on three benchmark datasets: MNIST \cite{lecun1998gradient}, FMNIST \cite{xiao2017fashion}, and CIFAR-10 \cite{krizhevsky2009learning}. MNIST/FMNIST contains 70,000 images of handwritten digits/clothing items from 10 different classes, with 60,000 for training and 10,000 for testing. CIFAR-10 consists of 60,000 RGB images from 10 different classes, with 50,000 for training and 10,000 for testing. Due to the different data sizes of these datasets, we set the total number of clients in MNIST and FMNIST to 10 and the total number of clients in CIFAR-10 to 20. To control the data heterogeneity relating to clients, we assign 5 of 10 labels for each client on MNIST, FMNIST, and CIFAR-10. 

\textbf{Baselines.} We compare BPFed with both typical PFL methods and the PFL methods with a neural or probabilistic shared-personalized structure. The former includes Per-FedAvg \cite{fallah2020personalized}, pFedMe \cite{t2020personalized}, and pFedBayes \cite{zhang2022personalized}. The latter includes FedPer \cite{arivazhagan2019federated}, LG-FedAvg \cite{liang2019think}, FedRep \cite{collins2021exploiting}, and FedSOUL \cite{kotelevskii2022fedpop}. In addition, we also compare BPFed with FedAvg \cite{mcmahan2017communication} as it is a commonly used FL method.

\textbf{Model settings.} We set the random set of clients $S$ to 10 for all the experiments on MNIST, FMNIST, and CIFAR-10. We perform 10 epochs to train the local model for each client. In addition, we follow the model settings in \cite{zhang2022personalized} and \cite{collins2021exploiting} using 1-layer multilayer perceptron for MNIST and FMNIST and 5-layer convolutional neural networks (2 convolutional layers followed by 3 fully-connected layers) for CIFAR-10. In BPFed, we arrange the parameters of the last layer as the personalized factors. We run all experiments in this paper on a system comprising 71 cores, with each core powered by an Intel(R) Xeon(R) Gold 6354 CPU operating at a frequency of 3.00GHz. The system was further enhanced by the inclusion of four NVIDIA Tesla A800 GPUs, each equipped with 80GB of memory. In addition to the GPU memory, the system boasted a total of 500GB of memory.

\textbf{Hyperparameters.} 
In our experiments, we conducted a search for the optimal learning rate within the set of values \{1e-3, 5e-3, 1e-2, 5e-2, 0.1\}. Through this search, we found that a learning rate of 1e-3 yielded the best results for most algorithms. Therefore, we selected 1e-3 as the common learning rate for all algorithms on each dataset. Specifically, for pFedMe, we chose a learning rate of 1e-2. Additionally, we set the batch size to 50 for all algorithms. Regarding the optimizer, we utilized the Adam optimizer \cite{kingma2014adam} to train our model and most baseline algorithms. For pFedMe and Per-FedAvg, we employed the optimizers introduced in their respective papers.

\subsection{Effect of data size on classification}
\label{6.2}

\begin{figure}[!t]
  \centering
  \begin{subfigure}{0.24\textwidth}
  \includegraphics[width=\textwidth]{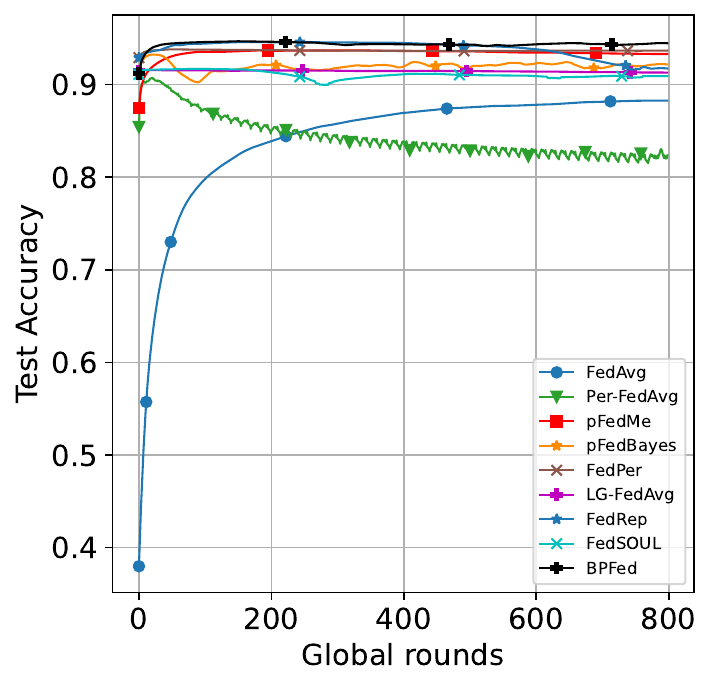}
  %\caption{Small}
  \end{subfigure}
  \begin{subfigure}{0.24\textwidth}
  \includegraphics[width=\textwidth]{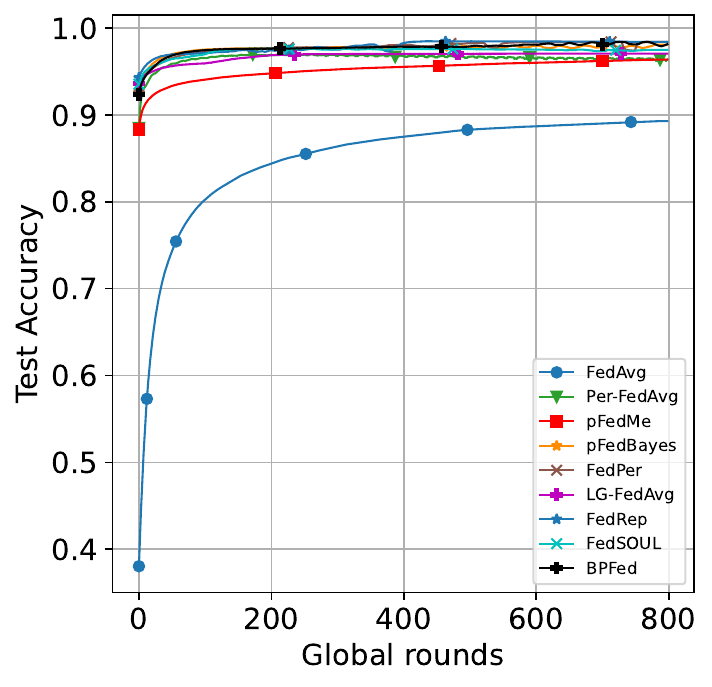}
  %\caption{Large}
  \end{subfigure}
  \label{fig:subfig6}
  \vspace{-2mm}
  \caption{Convergence results of different algorithms on small (left) and large (right) subsets of MNIST.}
  \label{figure 3}
  \vspace{-4mm}
\end{figure}

\begin{figure}[!t]
  \centering
  \begin{subfigure}{0.24\textwidth}
  \includegraphics[width=\textwidth]{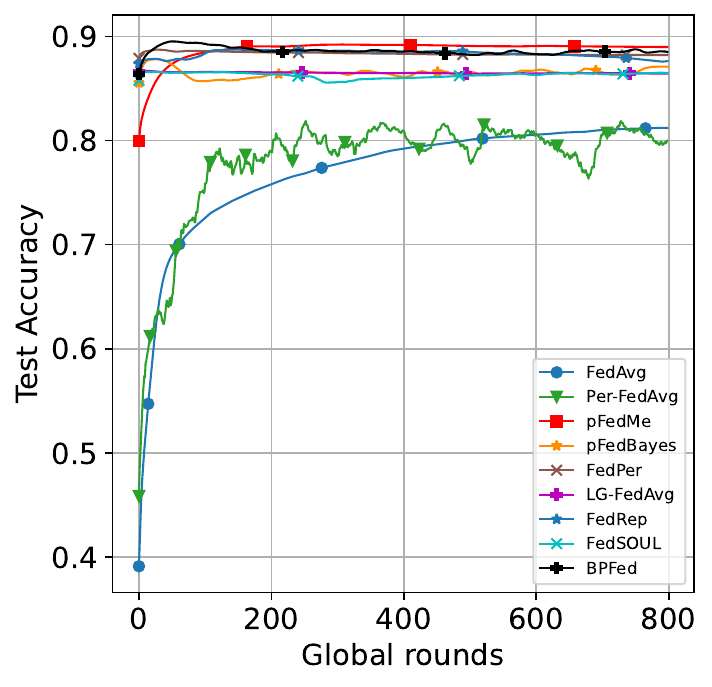}
  %\caption{Small}
  \end{subfigure}
  \begin{subfigure}{0.24\textwidth}
  \includegraphics[width=\textwidth]{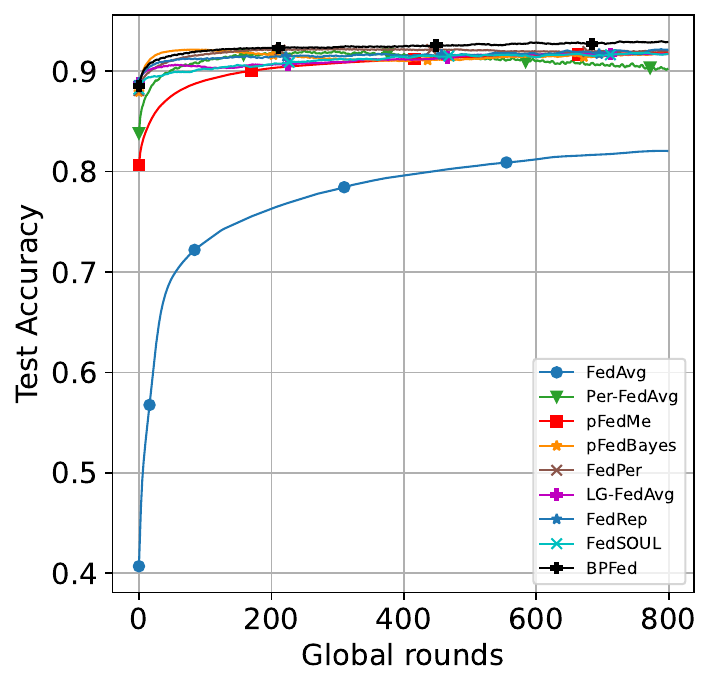}
  %\caption{Large}
  \end{subfigure}
  % \label{fig:subfig6}
  \vspace{-2mm}
  \caption{Convergence results of different algorithms on small (left) and large (right) subsets of FMNIST.}
  \label{figure 4}
  \vspace{-4mm}
\end{figure}

\begin{figure}[!t]
  \centering
  \begin{subfigure}{0.24\textwidth}
  \includegraphics[width=\textwidth]{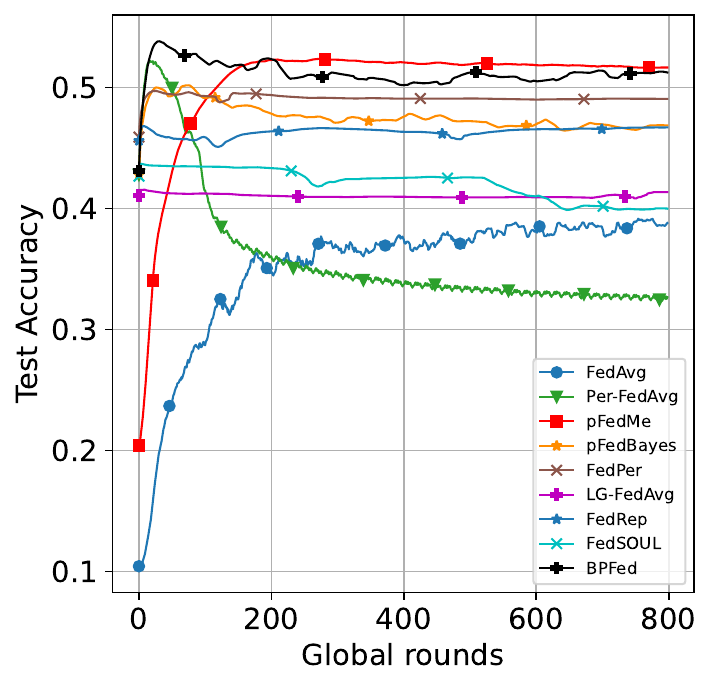}
  %\caption{Small}
  \end{subfigure}
  \begin{subfigure}{0.24\textwidth}
  \includegraphics[width=\textwidth]{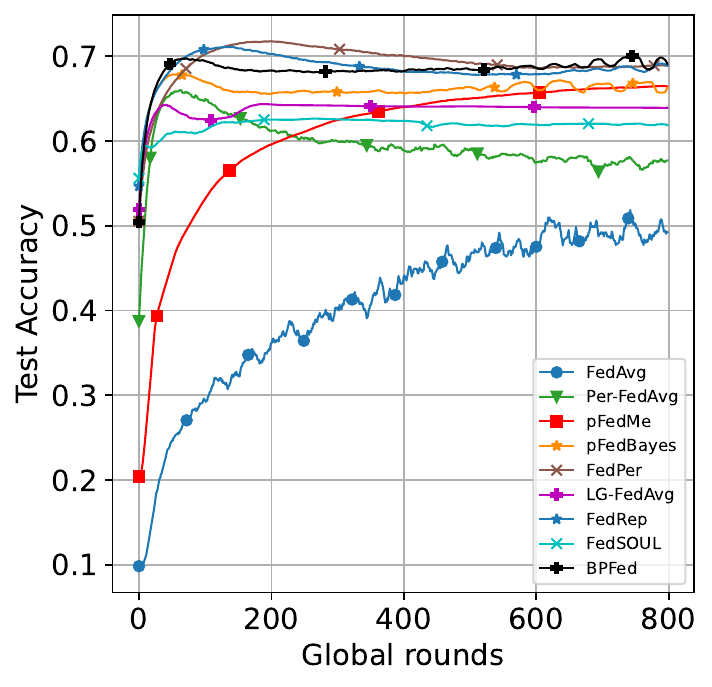}
  %\caption{Large}
  \end{subfigure}
  % \label{fig:subfig6}
  \vspace{-2mm}
  \caption{Convergence results of different algorithms on small (left) and large (right) subsets of CIFAR-10.}
  \label{figure 5}
  \vspace{-4mm}
\end{figure}

By introducing an appropriate prior distribution, our method can prevent the overfitting of local models on limited data. We test this capacity by splitting each dataset into small and large subsets following the data partition strategy in \cite{zhang2022personalized}. Specifically, to create small and large subsets of MNIST and FMNIST, we allocate 50/900 samples in training and 950/300 samples in test for each class respectively. For CIFAR-10, we allocate 25/450 samples in training and 475/150 samples in test for each class in the  small and large datasets respectively. We evaluate the modeling performance w.r.t. the highest accuracy over entire communication rounds. The numerical results are shown in Table \ref{table 1}, BPFed outperforms the baselines on all datasets except the large subsets of CIFAR-10. On the large subsets of CIFAR-10, FedPer and FedRep, which have similar model structures to BPFed, perform better than BPFed. This may be related to various aspects, such as mean-field assumption, etc. Figs. \ref{figure 3}-\ref{figure 5} show the convergence results of different algorithms on all datasets. We can observe that: (1) BPFed yields relatively stable convergence results over all different configurations. (2) On three small subsets of MNIST, FMNIST and CIFAR-10, BPFed achieves the fastest convergence rate and the highest test accuracy.

%\begin{figure}[!t]
  %\centering
  %\subfloat
  %{\includegraphics[width=0.25\textwidth]{Figures/Classfication/MNISTSMALL.pdf}\label{fig:subfig5}}
  %\subfloat
  %{\includegraphics[width=0.25\textwidth]{Figures/Classfication/MNISTLARGE.pdf}\label{fig:subfig6}}
  %\caption{Convergence results of different algorithms on small (left) and large (right) subsets of MNIST.}
  %\label{figure 3}
%\end{figure}

%\begin{figure}[!t]
  %\centering
   %\subfloat
  %{\includegraphics[width=0.25\textwidth]{Figures/Classfication/FMNISTSMALL.pdf}\label{fig:subfig7}}
  %\subfloat
   %{\includegraphics[width=0.25\textwidth]{Figures/Classfication/FMNISTLARGE.pdf}\label{fig:subfig8}}
  %\caption{Convergence results of different algorithms on small (left) and large (right) subsets of FMNIST.}
  %\label{figure 4}
%\end{figure}

%\begin{figure}[!t]
  %\centering
   %\subfloat
  %{\includegraphics[width=0.25\textwidth]{Figures/Classfication/CIFAR10SMALL.pdf}\label{fig:subfig9}}
  %\subfloat
   %{\includegraphics[width=0.25\textwidth]{Figures/Classfication/CIFAR10LARGE.pdf}\label{fig:subfig10}}
  %\caption{Convergence results of different algorithms on small (left) and large (right) subsets of CIFAR-10.}
  %\label{figure 5}
%\end{figure}

\subsection{Effect of uncertainty quantification}
\label{6.1}
\begin{figure}[!t]
  \centering
  \begin{subfigure}{0.24\textwidth}
  \includegraphics[width=\textwidth]{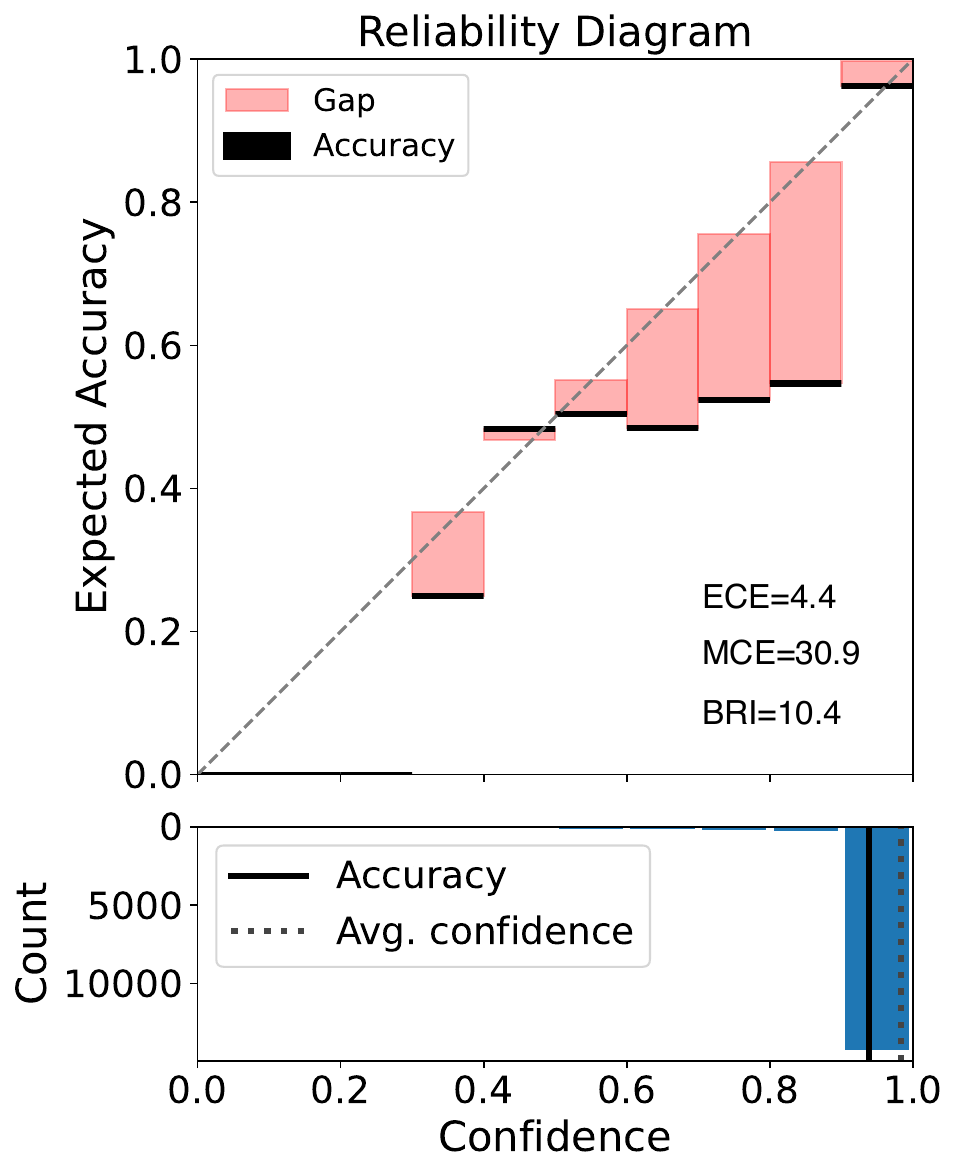}
  %\caption{Small}
  \end{subfigure}
  \begin{subfigure}{0.24\textwidth}
  \includegraphics[width=\textwidth]{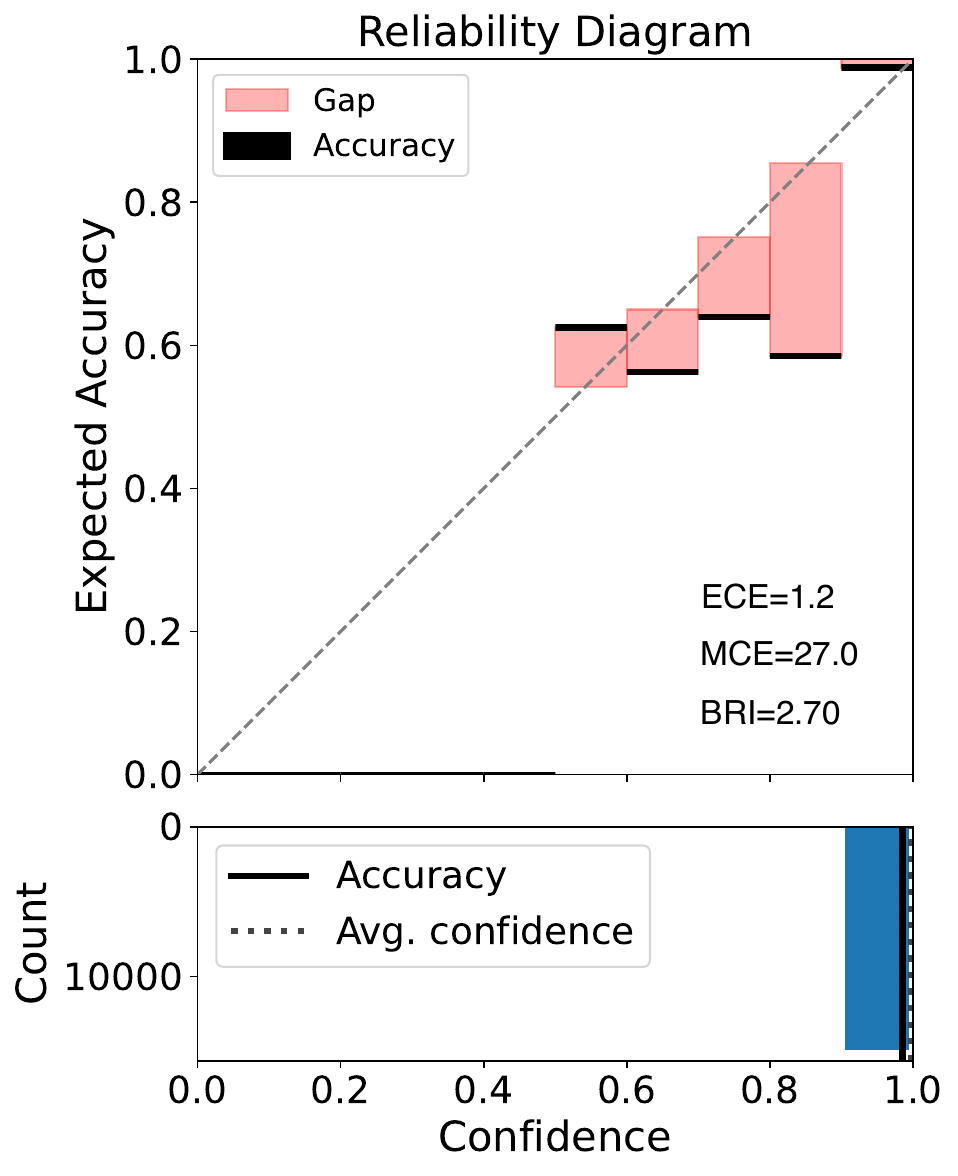}
  %\caption{Large}
  \end{subfigure}
  % \label{fig:subfig6}
  \vspace{-2mm}
  \caption{Reliability diagram (top) and confidence histogram (bottom) of BPFed on small (left) and large (right) subsets of MNIST. The closer the accuracy line and the average confidence line are, the better the model is calibrated.}
  \label{figure 6}
  \vspace{-5mm}
\end{figure}

As BPFed aims to mitigate existing gaps by quantifying uncertainty, its model parameters follow a Gaussian distribution rather than fitted per data points. 
%which is a desired property for medical and safety-critical areas (\cite{achituve2021personalized,yang2021toward}). 
Therefore, here, we evaluate this effect by conducting confidence calibration experiments for BPFed \cite{guo2017calibration,minderer2021revisiting}. We analyze this property in terms of visual representation and related metrics: expected calibration error (ECE), maximum calibration error (MCE), and brier score (BRI). 

%\begin{figure}[!t]
  %\centering
  %\subfloat
  %{\includegraphics[width=0.25\textwidth]{Figures/Uncertainty quantification/Mnist_small_BPFedPD_p_0.001_1.0_15_10u_100b_20_0.pdf}\label{fig:subfig5}}
  %\subfloat
  %{\includegraphics[width=0.25\textwidth]{Figures/Uncertainty quantification/Mnist_large_BPFedPD_p_0.001_1.0_15_10u_100b_20_0.pdf}\label{fig:subfig6}}
  %\caption{Reliability diagram (top) and confidence histogram (bottom) of BPFed on small (left) and large (right) subsets of MNIST. The closer the accuracy line and the average confidence line are, the better the model is calibrated.}
  %\label{figure 6}
%\end{figure}

Fig. \ref{figure 6} shows the calibration results on small and large MNIST subsets. It shows that BPFed is well-calibrated on both datasets. Table \ref{table 2} shows the comparison results of expected calibration error between BPFed and two other Bayesian FL methods, pFedBayes and FedSOUL, on MNIST, FMNIST and CIFAR-10 datasets, we can observe that BPFed achieves improved confidence calibration results compared to the two other baselines. This proves the superiority of BPFed compared to other Bayesian FL methods in terms of uncertainty quantification performance.

\begin{table}[!t]
\renewcommand{\arraystretch}{1.3}
\caption{Comparison of expected calibration error (ECE)  on MNIST, FMNIST and CIFAR-10 w.r.t. small and large size.}
\label{table 2}
\centering
\begin{tabular}{ccccccc}
\toprule
& \multicolumn{2}{c}{MNIST} & \multicolumn{2}{c}{FMNIST} & \multicolumn{2}{c}{CIFAR-10} \\
 \midrule 
Dataset size                              & Small & Large & Small & Large & Small & Large \\
 \midrule

 pFedBayes   & 4.7 & 1.4 & 9.2 & 7.1 & 46.3 & 35.2  \\
 FedSOUL   & 6.2 & 1.9 & 9.7 & 7.8 & 52.0 & 43.3 \\
\midrule 
BPFed (Ours)                              & $\boldsymbol{4.4}$& $\boldsymbol{1.2}$& $\boldsymbol{8.9}$ & $\boldsymbol{6.6}$ & $\boldsymbol{42.8}$ & $\boldsymbol{29.8}$ \\
\bottomrule
\end{tabular}
\end{table} 

\subsection{Generalization to novel clients}
\label{6.3}

\begin{figure}[!t]
  \centering
  \begin{subfigure}[small]{0.24\textwidth}
  \includegraphics[width=\textwidth]{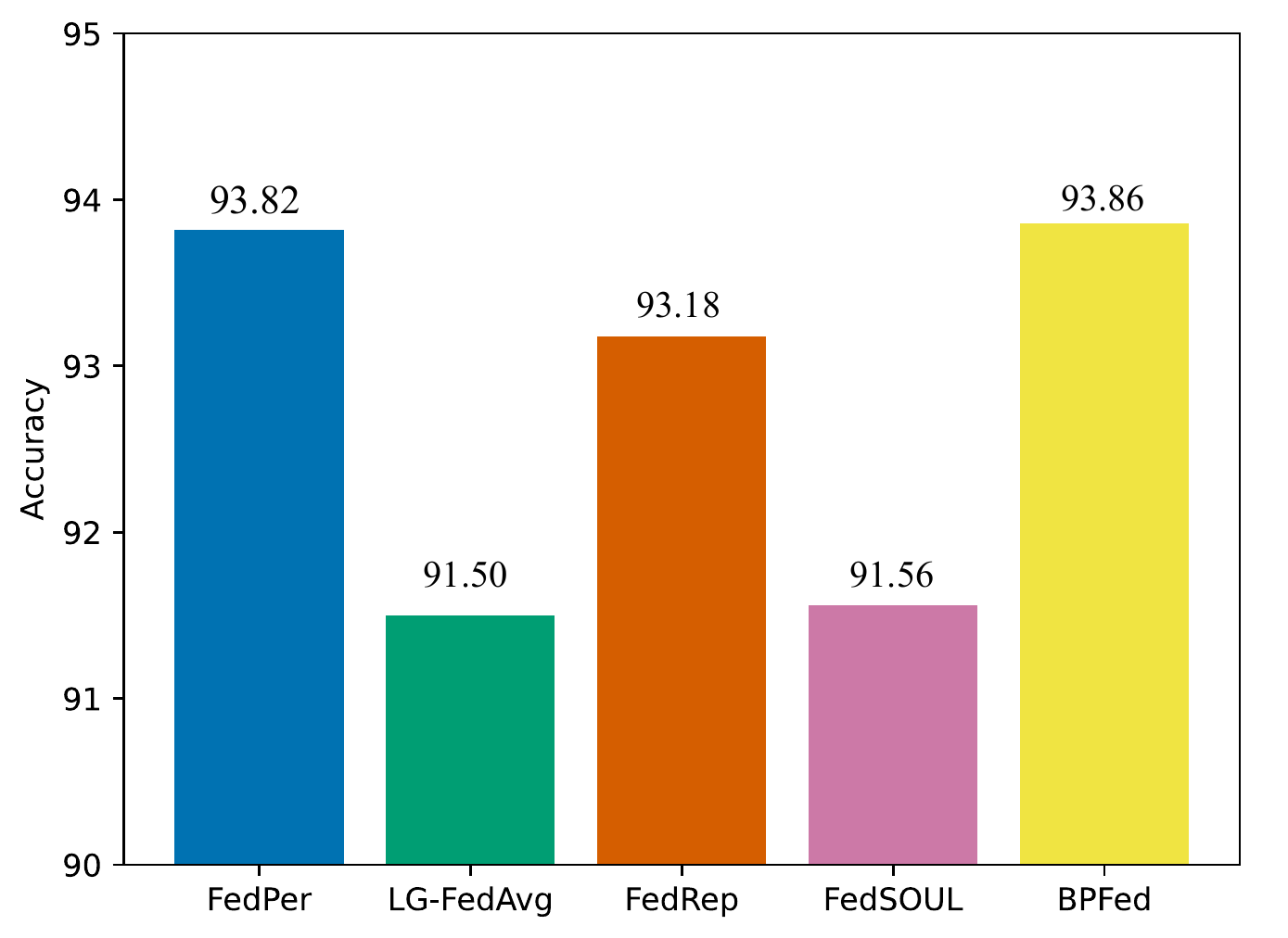}
  %\caption{Small}
  \end{subfigure}
  \begin{subfigure}[large]{0.24\textwidth}
  \includegraphics[width=\textwidth]{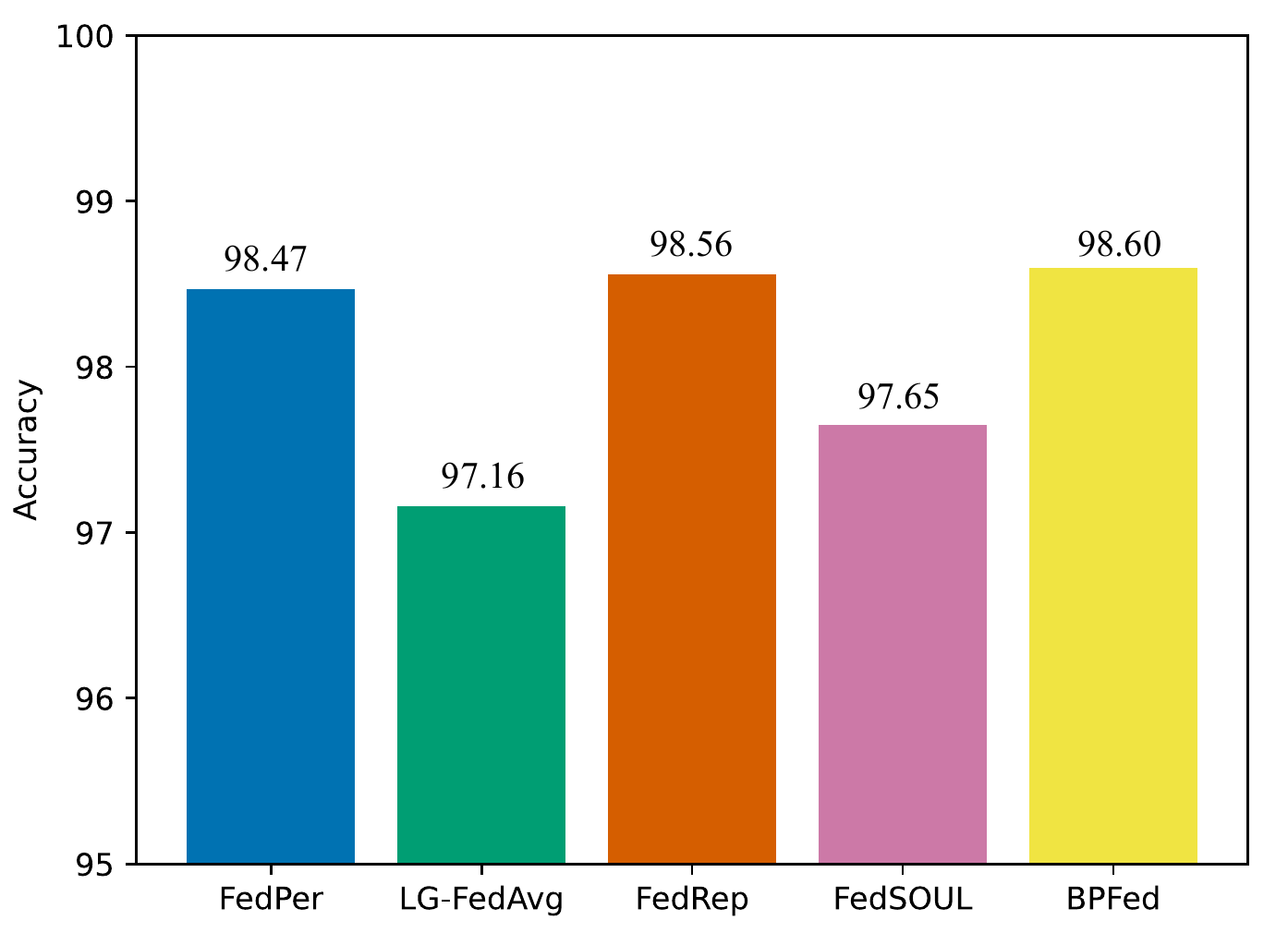}
  %\caption{Large}
  \end{subfigure}
  % \label{fig:subfig6}
  %\vspace{-2mm}
  \caption{Test accuracy of different algorithms on small (left) and large (right) subsets of MNIST.}
  \label{figure 7}
  \vspace{-4mm}
\end{figure}

\begin{figure}[!t]
  \centering
  \begin{subfigure}[small]{0.24\textwidth}
  \includegraphics[width=\textwidth]{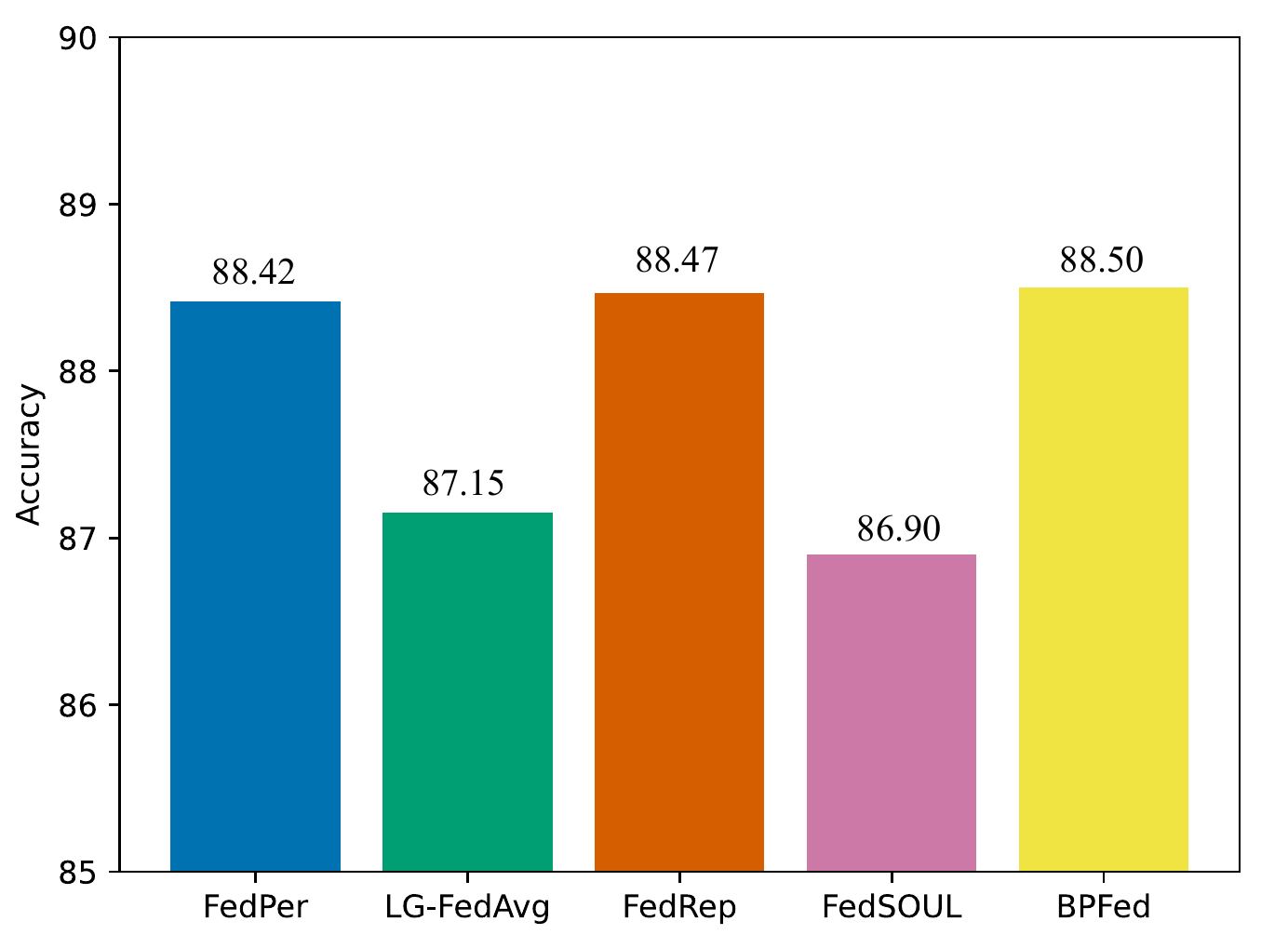}
  %\caption{Small}
  \end{subfigure}
  \begin{subfigure}[large]{0.24\textwidth}
  \includegraphics[width=\textwidth]{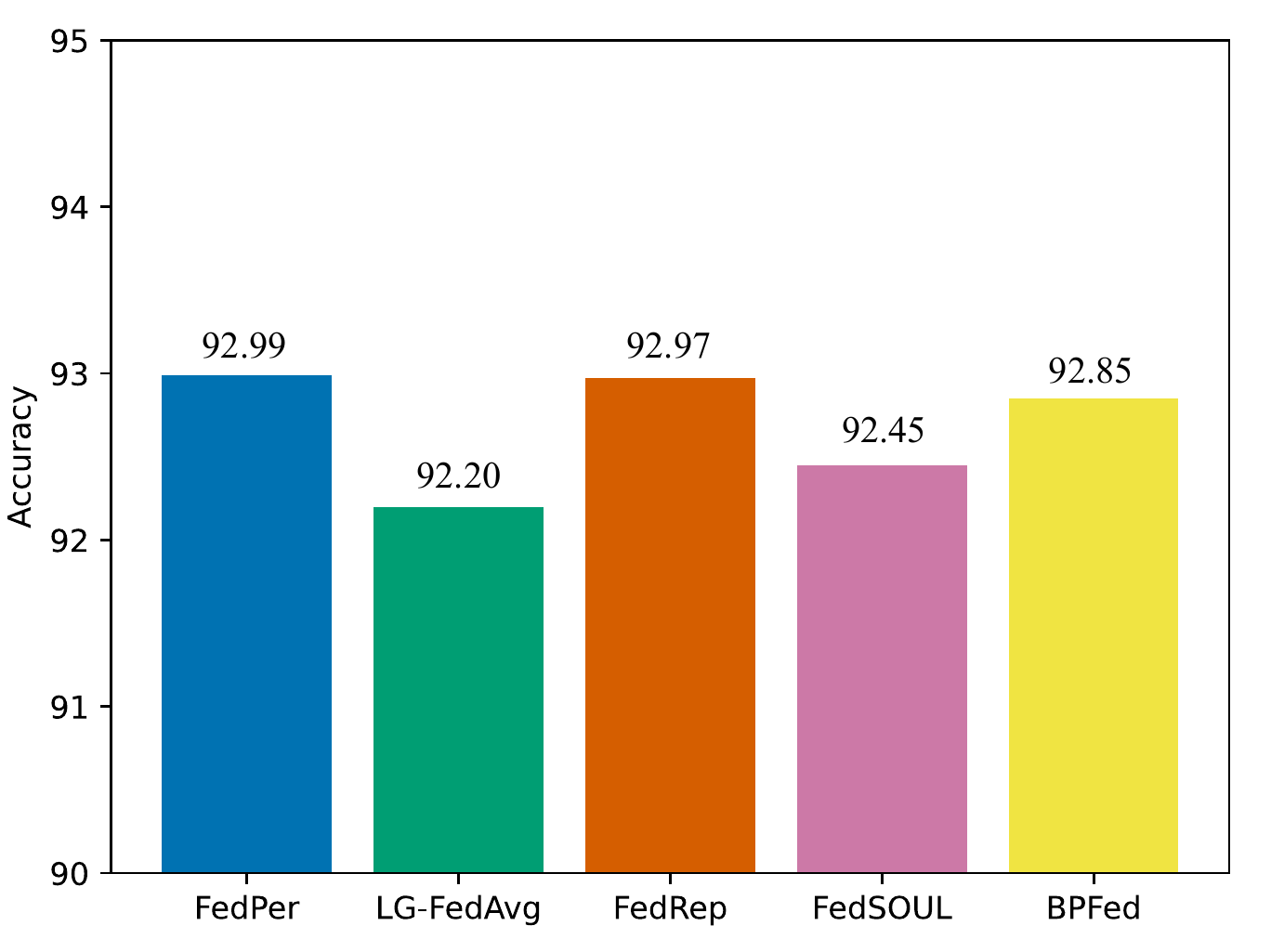}
  %\caption{Large}
  \end{subfigure}
  %\vspace{-2mm}
  \caption{Test accuracy of different algorithms on small (left) and large (right) subsets of FMNIST.}
  \label{figure 8}
  \vspace{-4mm}
\end{figure}

\begin{figure}[!t]
  \centering
  \begin{subfigure}[small]{0.24\textwidth}
  \includegraphics[width=\textwidth]{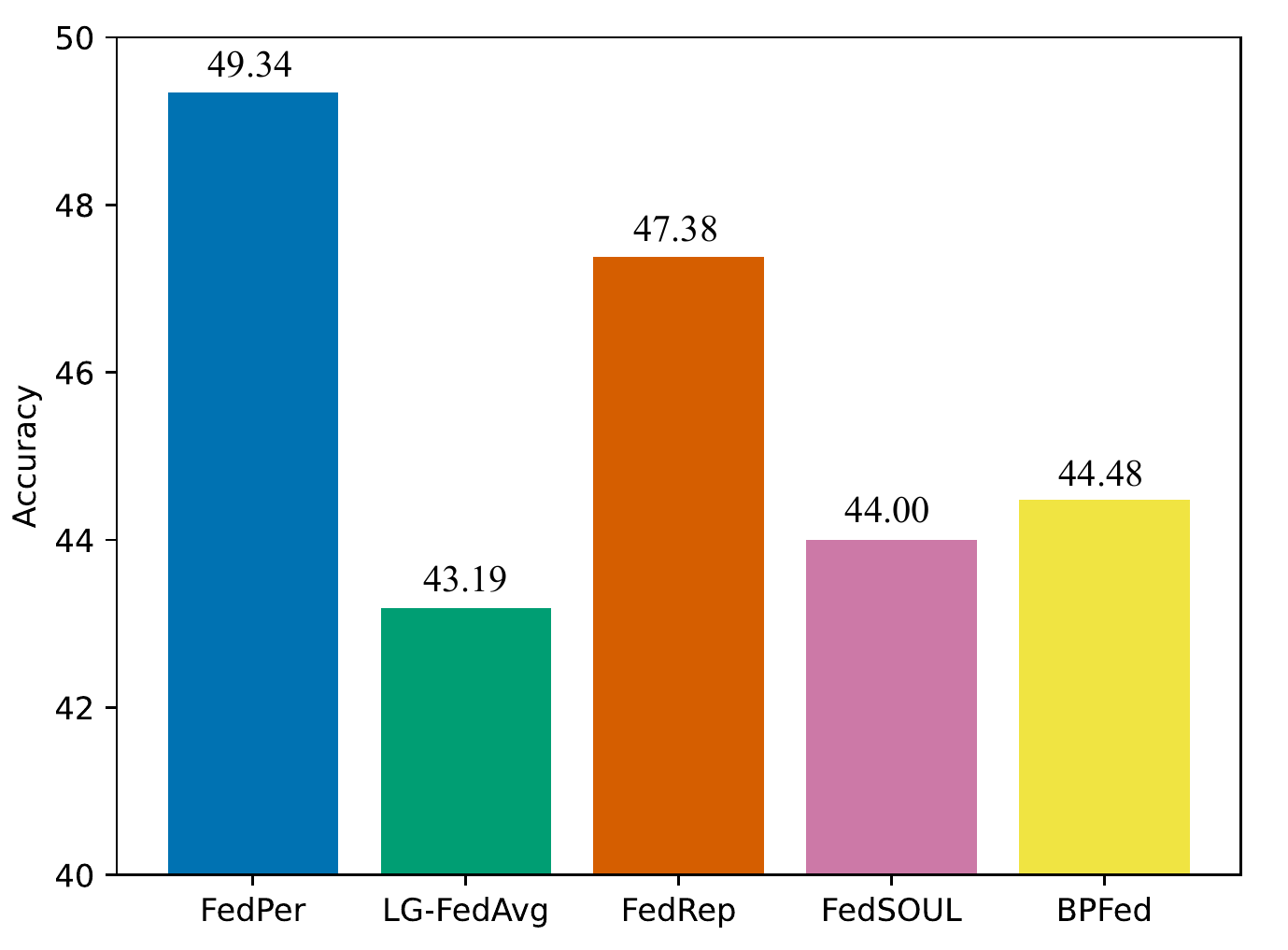}
  %\caption{Small}
  \end{subfigure}
  \begin{subfigure}[large]{0.24\textwidth}
  \includegraphics[width=\textwidth]{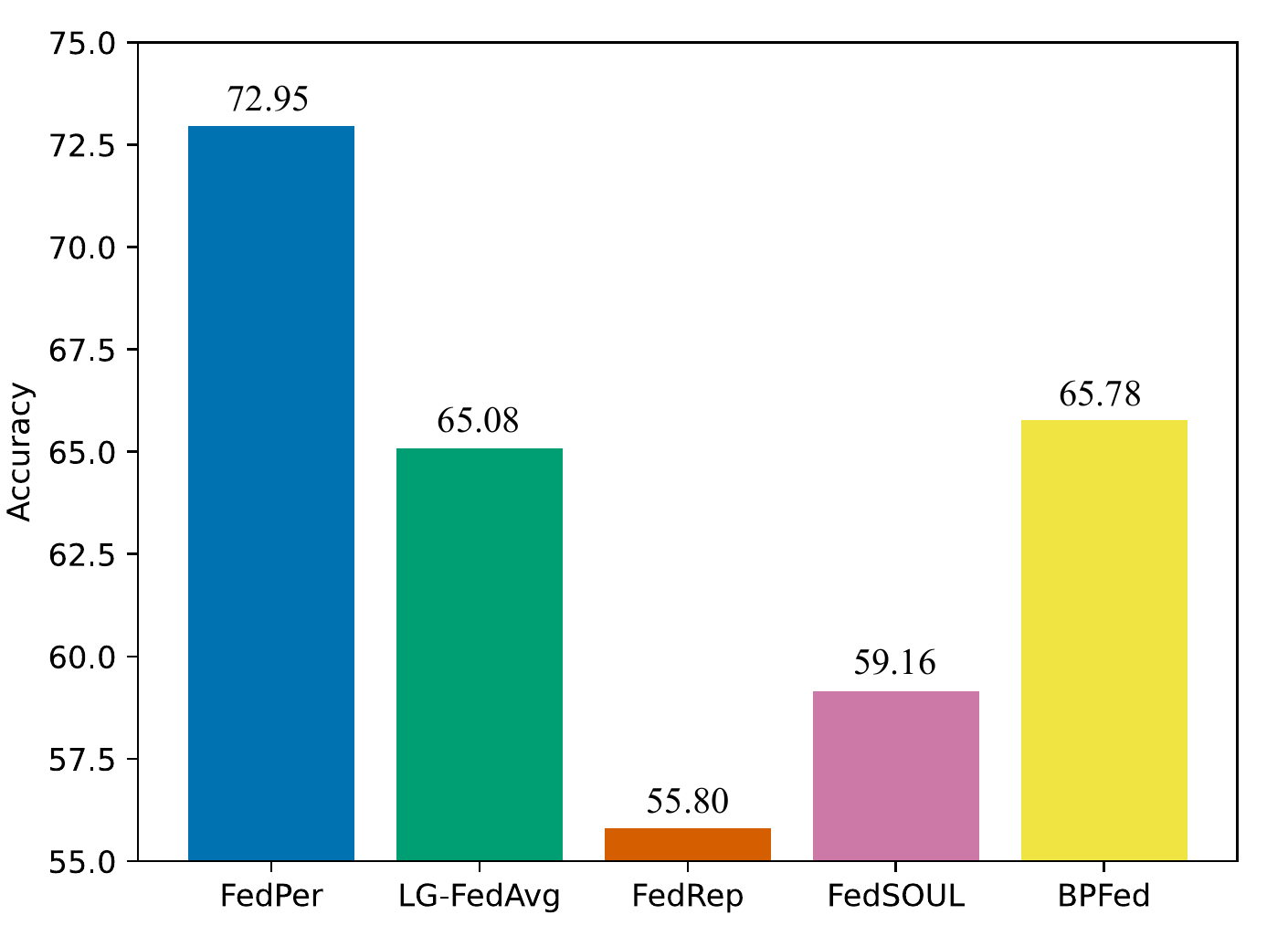}
  %\caption{Large}
  \end{subfigure}
  %\vspace{-2mm}
  \caption{Test accuracy of different algorithms on small (left) and large (right) subsets of CIFAR-10.}
  \label{figure 9}
  \vspace{-4mm}
\end{figure}

Since BPFed is a dynamic learning system,  we evaluate its capacity in handling novel clients when unseen clients participate in the tasks. The model with shared-personalized structure have a significant advantage for this due to the shared factors. Therefore, we follow the similar experimental protocol as in \cite{collins2021exploiting} and choose all PFL methods with shared-personalized structure as our baselines. For each dataset, we divide clients into two parts: a novel client for model test and other clients for model training. The model training obtains the shared factors, which are combined with the samples of novel client as input to learn low-dimensional personalized factors. This way incurs low computational costs. Figs. \ref{figure 7}-\ref{figure 9} show the average accuracy of these methods on the all datasets.  We can find that BPFed achieves the highest accuracy on small and large subsets of MNIST. On the small and large subsets of FMNIST, BPFed outperforms LG-FedAvg and FedSOUL by a large margin, but on par with FedPer and FedRep. Regarding the results from CIFAR-10 dataset, we notice that FedPer have the best performance and BPFed still performs better than LG-FedAvg and FedSOUL on both subsets; Surprisingly, FedRep performs exceptionally well on the small subset of CIFAR-10, only trailing behind FedPer, but it exhibits the poorest performance on the large subset of CIFAR-10.

\section{Conclusions}

Non-IID real-life federated systems involve client heterogeneity and uncertainty. We address these issues by novel Bayesian personalized federated learning with statistical heterogeneity and uncertainty. In a Bayesian approach, a general factor decomposition framework BPFed is proposed to decouple shared factors from personalized factors characterizing each client. It estimates the shared uncertainty and personalized uncertainty of a client over communications between clients and the server. The shared and personalized uncertainty learning is guaranteed by a generalization error bound with continual prior updating over iterations. BPFed has the potential in quantifying uncertainties relating to heterogeneous, novel and dynamic clients by jointly decomposing and coupling both dependent (shared) and independent (personalized) factors. 

%\section*{Acknowledgments}
%This should be a simple paragraph before the References to thank those individuals and institutions who have supported your work on this article.
\appendix
\section*{Proof of the Lemma 1}
To prove the Lemma 1, we begin by defining the optimal values for $\boldsymbol{\phi}_i$ and $\boldsymbol{\theta}$:
\begin{equation}
(\boldsymbol{\phi}_i^*, \boldsymbol{\theta}^*)=\arg\min_{\boldsymbol{\phi}_i, \boldsymbol{\theta} \in \Theta\left(L, \boldsymbol{e}\right),\|\boldsymbol{\phi}_i, \boldsymbol{\theta}\|_{\infty} \leq B}\left\|f_{\boldsymbol{\phi}_i, \boldsymbol{\theta}}-f_i\right\|_{\infty}^2, 
\end{equation}
we then denote the optimal solution $\hat{q}(\boldsymbol{\phi}_i, \boldsymbol{\theta})$ as follows, for $t_1=1,\dots,T_1$, $t_2=1,\dots,T_2$:
\begin{equation}
 \phi_{i,t_1}\sim \mathcal{N}(\phi_{i,t_1}^*,\sigma^2),
\end{equation}
\begin{equation}
 \theta_{i,t_2}\sim \mathcal{N}(\theta_{i,t_2}^*,\sigma^2),
\end{equation}
where we have 
\begin{equation}
\begin{aligned}
&\sigma^2=\frac{T_1+T_2}{8n} \log ^{-1}(3 e_0 K)(2 B K)^{-2(L+1)}\\
&\left\{\left(e_0+\frac{BK}{B K-1}\right)^2+\frac{1}{(2 B K)^2-1}+\frac{2}{(2 B K-1)^2}\right\}^{-1}.
\end{aligned}
\end{equation}

Next, we define $\pi^*(\boldsymbol{\phi}_i, \boldsymbol{\theta})$ as the optimal prior distribution for $i$-th client, thus
\begin{equation}
\pi^*(\boldsymbol{\phi}_i, \boldsymbol{\theta})=\arg\min _{\pi(\boldsymbol{\phi}_i, \boldsymbol{\theta})} \frac{1}{N} \sum_{i=1}^N \mathrm{KL}\left(\hat{q}(\boldsymbol{\phi}_i, \boldsymbol{\theta}) \| \pi(\boldsymbol{\phi}_i, \boldsymbol{\theta})\right) .
\end{equation}

To obtain the distribution parameters $\mu_{{\eta}_{\pi},t_1}^*,\sigma_{{\eta}_{\pi},t_1}^*$, $\mu_{{\zeta}_{\pi},t_2}^*$ and $\sigma_{{\zeta}_{\pi},t_2}^*$ of $\pi^*(\boldsymbol{\phi}_i, \boldsymbol{\theta})$, we have following formulations:
\begin{equation}
\label{eq.13}
\frac{1}{N} \sum_{i=1}^N \frac{\partial \mathrm{KL}\left(\hat{q}(\boldsymbol{\phi}_i, \boldsymbol{\theta}) \| \pi(\boldsymbol{\phi}_i, \boldsymbol{\theta})\right)}{\mu_{{\eta}_\pi,t_1}}=0,
\end{equation}
\begin{equation}
\label{eq.14}
\frac{1}{N} \sum_{i=1}^N \frac{\partial \mathrm{KL}\left(\hat{q}(\boldsymbol{\phi}_i, \boldsymbol{\theta}) \| \pi(\boldsymbol{\phi}_i, \boldsymbol{\theta})\right)}{\sigma_{{\eta}_\pi,t_1}}=0,
\end{equation}
\begin{equation}
\label{eq.15}
\frac{1}{N} \sum_{i=1}^N \frac{\partial \mathrm{KL}\left(\hat{q}(\boldsymbol{\phi}_i, \boldsymbol{\theta}) \| \pi(\boldsymbol{\phi}_i, \boldsymbol{\theta})\right)}{\mu_{{\zeta}_\pi,t_2}}=0,
\end{equation}
\begin{equation}
\label{eq.16}
\frac{1}{N} \sum_{i=1}^N \frac{\partial \mathrm{KL}\left(\hat{q}(\boldsymbol{\phi}_i, \boldsymbol{\theta}) \| \pi(\boldsymbol{\phi}_i, \boldsymbol{\theta})\right)}{\sigma_{{\zeta}_\pi,t_2}}=0.
\end{equation}

Furthermore, since
\begin{equation}
\begin{aligned}
&\mathrm{KL}\left(q(\boldsymbol{\phi}_i, \boldsymbol{\theta}) \| \pi(\boldsymbol{\phi}_i, \boldsymbol{\theta})\right)\\
= & \operatorname{KL}\left(\prod_{t_1=1}^{T_1} \mathcal{N}\left(\mu_{\eta_i, t_1}, \sigma_{{\eta}_i,t_1}^2\right) \prod_{t_2=1}^{T_2} \mathcal{N}\left(\mu_{\zeta_i, t_2}, \sigma_{{\zeta}_i,t_2}^2\right) \| \right.\\
&\left.\prod_{t_1=1}^{T_1} \mathcal{N}\left(\mu_{\eta_{\pi}, t_1}, \sigma_{\eta_{\pi}, t_1}^2\right)\prod_{t_2=1}^{T_2} \mathcal{N}\left(\mu_{\zeta_\pi, t_2}, \sigma_{{\zeta}_\pi,t_2}^2\right)\right) \\
= & \sum_{t_1=1}^{T_1} \operatorname{KL}\left(\mathcal{N}\left(\mu_{\eta_i, t_1}, \sigma_{{\eta}_i,t_1}^2\right) \| \mathcal{N}\left(\mu_{\eta_{\pi}, t_1}, \sigma_{\eta_{\pi}, t_1}^2\right)\right)\\ 
&+ \sum_{t_2=1}^{T_2} \operatorname{KL}\left(\mathcal{N}\left(\mu_{\zeta_i, t_2}, \sigma_{{\zeta}_i,t_2}^2\right) \| \mathcal{N}\left(\mu_{\zeta_\pi, t_2}, \sigma_{{\zeta}_\pi,t_2}^2\right)\right)\\
= & \frac{1}{2} \sum_{t_1=1}^{T_1}\left[\log \left(\frac{\sigma_{\eta_{\pi}, t_1}^2}{\sigma_{{\eta}_i,t_1}^2}\right)+\frac{\sigma_{{\eta}_i,t_1}^2+\left(\mu_{\eta_i, t_1}-\mu_{\eta_{\pi}, t_1}\right)^2}{\sigma_{\eta_{\pi}, t_1}^2}-1\right] \\ 
&+ \frac{1}{2} \sum_{t_2=1}^{T_2}\left[\log \left(\frac{\sigma_{\zeta_{\pi}, t_2}^2}{\sigma_{{\zeta}_i,t_2}^2}\right)+\frac{\sigma_{{\zeta}_i,t_2}^2+\left(\mu_{\zeta_i, t_2}-\mu_{\zeta_{\pi}, t_2}\right)^2}{\sigma_{\zeta_{\pi}, t_2}^2}-1\right].
\end{aligned}
\end{equation}

Therefore, we have
\begin{equation}
\label{eq.18}
\begin{aligned}
\frac{1}{N} \sum_{i=1}^N \frac{\partial \mathrm{KL}\left(q(\boldsymbol{\phi}_i, \boldsymbol{\theta}) \| \pi(\boldsymbol{\phi}_i, \boldsymbol{\theta})\right)}{\mu_{{\eta}_\pi,t_1}}=\frac{1}{N} \sum_{i=1}^{N} \frac{\mu_{\eta_{\pi}, t_1}-\mu_{{\eta_{i},t_1} }}{\sigma_{\eta_{\pi}, t_1}^2},
\end{aligned}
\end{equation}

\begin{equation}
\begin{aligned}
\label{eq.19}
&\frac{1}{N} \sum_{i=1}^N \frac{\partial \mathrm{KL}\left(q(\boldsymbol{\phi}_i, \boldsymbol{\theta}) \| \pi(\boldsymbol{\phi}_i, \boldsymbol{\theta})\right)}{\sigma_{{\eta}_\pi,t_1}}\\
=&\frac{1}{N} \sum_{i=1}^{N} \left[\frac{1}{\sigma_{{\eta}_\pi,t_1}}-\frac{\sigma_{{\eta}_i,t_1}^2+\left(\mu_{{\eta}_i,t_1}-\mu_{{\eta}_\pi,t_1}\right)^2}{\sigma_{{\eta}_\pi,t_1}^3}\right],
\end{aligned}
\end{equation}

\begin{equation}
\label{eq.20}
\frac{1}{N} \sum_{i=1}^N \frac{\partial \mathrm{KL}\left(q(\boldsymbol{\phi}_i, \boldsymbol{\theta}) \| \pi(\boldsymbol{\phi}_i, \boldsymbol{\theta})\right)}{\mu_{{\zeta}_\pi,t_2}}=\frac{1}{N} \sum_{i=1}^{N} \frac{\mu_{\zeta_{\pi}, t_2}-\mu_{{\zeta_{i},t_2} }}{\sigma_{\zeta_{\pi}, t_2}^2},
\end{equation}
\begin{equation}
\begin{aligned}
\label{eq.21}
&\frac{1}{N} \sum_{i=1}^N \frac{\partial \mathrm{KL}\left(q(\boldsymbol{\phi}_i, \boldsymbol{\theta}) \| \pi(\boldsymbol{\phi}_i, \boldsymbol{\theta})\right)}{\sigma_{{\zeta}_\pi,t_2}}\\
=&\frac{1}{N} \sum_{i=1}^{N} \left[\frac{1}{\sigma_{{\zeta}_\pi,t_2}}-\frac{\sigma_{{\zeta}_i,t_2}^2+\left(\mu_{{\zeta}_i,t_2}-\mu_{{\zeta}_\pi,t_2}\right)^2}{\sigma_{{\zeta}_\pi,t_2}^3}\right] .
\end{aligned}
\end{equation}

To get the optimal solution, we let \eqref{eq.18}, \eqref{eq.19}, \eqref{eq.20} and \eqref{eq.21} equal zero, such that
\begin{equation}
\label{eq.22}
\mu_{{\eta}_\pi,t_1}  =\frac{1}{N} \sum_{i=1}^N \mu_{{\eta}_i,t_1},
\end{equation}
\begin{equation}
\label{eq.23}
\mu_{\zeta_{\pi}, t_2}  =\frac{1}{N} \sum_{i=1}^N \mu_{\zeta_i, t_2},
\end{equation}
\begin{equation}
\begin{aligned}
\label{eq.24}
\sigma_{{\eta}_\pi,t_1}^2  &=\frac{1}{N} \sum_{i=1}^N\left[\sigma_{{\eta}_i,t_1}^2+\left(\mu_{{\eta}_i,t_1}-\mu_{{\eta}_\pi,t_1}\right)^2\right]\\
&=\frac{1}{N} \sum_{i=1}^N\left[\sigma_{{\eta}_i,t_1}^2+\mu_{{\eta}_i,t_1}^2-\mu_{{\eta}_\pi,t_1}^2\right],
\end{aligned}
\end{equation}
\begin{equation}
\begin{aligned}
\label{eq.25}
\sigma_{{\zeta}_\pi,t_2}^2 &=\frac{1}{N} \sum_{i=1}^N\left[\sigma_{{\zeta}_i,t_2}^2+\left(\mu_{{\zeta}_i,t_2}-\mu_{{\zeta}_\pi,t_2}\right)^2\right]\\
&=\frac{1}{N} \sum_{i=1}^N\left[\sigma_{{\zeta}_i,t_2}^2+\mu_{{\zeta}_i,t_2}^2-\mu_{{\zeta}_\pi,t_2}^2\right].
\end{aligned}
\end{equation}

According to the \eqref{eq.22}, \eqref{eq.23}, \eqref{eq.24} and \eqref{eq.25}, then we can obtain
\begin{equation}
\label{eq.26}
\mu_{{\eta}_{\pi},t_1}^*  =\frac{1}{N} \sum_{i=1}^N \phi_{i,t_1}^*,
\end{equation}
\begin{equation}
\label{eq.27}
\mu_{{\zeta}_{\pi},t_2}^* =\frac{1}{N} \sum_{i=1}^N \theta_{i,t_2}^*,
\end{equation}
\begin{equation}
\begin{aligned}
\label{eq.28}
\sigma_{{\eta}_{\pi},t_1}^*  &=\frac{1}{N} \sum_{i=1}^N\left[\sigma^2+\left(\phi_{i,t_1}^*-\mu_{{\eta}_\pi,t_1}^*\right)^2\right]\\
&=\frac{1}{N} \sum_{i=1}^N \phi_{i,t_1}^{*2} + \sigma^2-\mu_{{\eta}_\pi,t_1}^{*2},
\end{aligned}
\end{equation}
\begin{equation}
\begin{aligned}
\label{eq.29}
\sigma_{{\zeta}_{\pi},t_2}^* &=\frac{1}{N} \sum_{i=1}^N\left[\sigma^2+\left(\theta_{i,t_2}^*-\mu_{{\zeta}_\pi,t_2}^*\right)^2\right]\\
&=\frac{1}{N} \sum_{i=1}^N \theta_{i,t_2}^{*2} + \sigma^2-\mu_{{\zeta}_\pi,t_2}^{*2}.
\end{aligned}
\end{equation}

Furthermore, to calculate the KL divergence between $\hat{q}(\boldsymbol{\phi}_i, \boldsymbol{\theta})$ and $\pi^*(\boldsymbol{\phi}_i, \boldsymbol{\theta})$, we have
\begin{equation}
\begin{aligned}
&\mathrm{KL}\left(\hat{q}(\boldsymbol{\phi}_i, \boldsymbol{\theta}) \| \pi^*(\boldsymbol{\phi}_i, \boldsymbol{\theta})\right)\\
= & \operatorname{KL}\left(\prod_{t_1=1}^{T_1} \mathcal{N}\left(\phi_{i,t_1}^*, \sigma^2\right) \prod_{t_2=1}^{T_2} \mathcal{N}\left(\theta_{i,t_2}^*, \sigma^2\right) \| \right.\\
&\left.\prod_{t_1=1}^{T_1} \mathcal{N}\left(\mu_{\eta_{\pi}, t_1}^*, \sigma_{\eta_{\pi}, t_1}^{*2}\right)\prod_{t_2=1}^{T_2} \mathcal{N}\left(\mu_{\zeta_\pi, t_2}^*, \sigma_{{\zeta}_\pi,t_2}^{*2}\right)\right) \\
= & \sum_{t_1=1}^{T_1} \operatorname{KL}\left(\mathcal{N}\left(\phi_{i,t_1}^*, \sigma^2\right) \| \mathcal{N}\left(\mu_{\eta_{\pi}, t_1}^*, \sigma_{\eta_{\pi}, t_1}^{*2}\right)\right)\\ 
&+ \sum_{t_2=1}^{T_2} \operatorname{KL}\left(\mathcal{N}\left(\theta_{i,t_2}^*, \sigma^2\right) \| \mathcal{N}\left(\mu_{\zeta_\pi, t_2}^*, \sigma_{{\zeta}_\pi,t_2}^{*2}\right)\right)\\
%= & \frac{1}{2} \sum_{t_1=1}^{T_1}\left[\log \left(\frac{\sigma_{\eta_{\pi}, t_1}^{*2}}{\sigma^2}\right)+\frac{\sigma^2+\left(\phi_{i,t_1}^*-\mu_{\eta_{\pi}, t_1}^*\right)^2}{\sigma_{\eta_{\pi}, t_1}^{*2}}-1\right] \\ 
%&+ \frac{1}{2} \sum_{t_2=1}^{T_2}\left[\log \left(\frac{\sigma_{\zeta_{\pi}, t_2}^{*2}}{\sigma^2}\right)+\frac{\sigma^2+\left(\theta_{i,t_2}^*-\mu_{\zeta_{\pi}, t_2}^*\right)^2}{\sigma_{\zeta_{\pi}, t_2}^{*2}}-1\right]\\
\leq & \frac{T_1+T_2}{2}\left[\log \left(\frac{\sigma^{2}+B^2}{\sigma^2}\right)-1\right]\\
&+\frac{1}{2} \sum_{t_1=1}^{T_1} \frac{\sigma^2+\left(\phi_{i,t_1}^*-\mu_{\eta_{\pi}, t_1}^*\right)^2}{\sigma_{\eta_{\pi}, t_1}^{*2}}\\
&+ \frac{1}{2} \sum_{t_2=1}^{T_2}\frac{\sigma^2+\left(\theta_{i,t_2}^*-\mu_{\zeta_{\pi}, t_2}^*\right)^2}{\sigma_{\zeta_{\pi}, t_2}^{*2}},
\end{aligned}
\end{equation}
where the inequality is due to \eqref{eq.28}, \eqref{eq.29} and Assumption 3.

Noting that
\begin{equation}
\frac{1}{N} \sum_{i=1}^N  \frac{\sigma^2+\left(\phi_{i,t_1}^*-\mu_{\eta_{\pi}, t_1}^*\right)^2}{\sigma_{\eta_{\pi}, t_1}^{*2}}=1  
\end{equation}
\begin{equation}
\frac{1}{N} \sum_{i=1}^N  \frac{\sigma^2+\left(\theta_{i,t_2}^*-\mu_{\zeta_{\pi}, t_2}^*\right)^2}{\sigma_{\zeta_{\pi}, t_2}^{*2}}=1  
\end{equation}

Then we have
\begin{equation}
\begin{aligned}
&\frac{1}{N} \sum_{i=1}^N \mathrm{KL}\left(\hat{q}(\boldsymbol{\phi}_i, \boldsymbol{\theta}) \| \pi^*(\boldsymbol{\phi}_i, \boldsymbol{\theta})\right)\\
\leq & \frac{1}{N} \sum_{i=1}^N\left\{\frac{T_1}{2}\left[\log \left(\frac{\sigma^{2}+B^2}{\sigma^2}\right)-1\right]\right.\\
&\left.+\frac{1}{2} \sum_{t_1=1}^{T_1} \frac{\sigma^2+\left(\phi_{i,t_1}^*-\mu_{\eta_{\pi}, t_1}^*\right)^2}{\sigma_{\eta_{\pi}, t_1}^{*2}}\right\} \\ 
&+ \frac{1}{N} \sum_{i=1}^N\left \{\frac{T_2}{2}\left[\log \left(\frac{\sigma^{2}+B^2}{\sigma^2}\right)-1\right]\right.\\
&+ \frac{1}{2} \sum_{t_2=1}^{T_2}\frac{\sigma^2+\left(\theta_{i,t_2}^*-\mu_{\zeta_{\pi} t_2}^*\right)^2}{\sigma_{\zeta_{\pi}, t_2}^{*2}}\bigg\}\\
= & \frac{T_1+T_2}{2} \log \left(\frac{\sigma^{2}+B^2}{\sigma^2}\right) \\
\leq &\frac{T_1+T_2}{2} \log \left(\frac{2 B^2}{\sigma_n^2}\right).
\end{aligned}
\end{equation}

Recall the definition of $\sigma^2$, we have
\begin{align}
&\frac{T_1+T_2}{2} \log \left(\frac{2 B^2}{\sigma_n^2}\right) \\
\leq &(T_1+T_2)(L+1) \log (2 B M) \nonumber\\
&+\frac{T}{2} \log \log \left(3 s_0 M\right)+T \log \left(4 s_0 \sqrt{\frac{n}{T}}\right)+\frac{T}{2} \log \left(2 B^2\right)\\ 
\leq &n C r^n. \nonumber
\end{align}

Therefore, we can obtain
\begin{equation}
\label{eq.35}
\frac{1}{N} \sum_{i=1}^N \mathrm{KL}\left(\hat{q}(\boldsymbol{\phi}_i, \boldsymbol{\theta}) \| \pi^*(\boldsymbol{\phi}_i, \boldsymbol{\theta})\right)\leq n C r^n.
\end{equation}

Furthermore, by the supplementary of \cite{bai2020efficient}, it shows 
\begin{equation}
\begin{aligned}
&\int_{\Theta}\int_{\Theta} \left\|f_{\boldsymbol{\phi}_i, \boldsymbol{\theta}}-f_{\boldsymbol{\phi}_i^*, \boldsymbol{\theta}^*}\right\|_{\infty}^2\hat{q}(\boldsymbol{\phi}_i,\boldsymbol{\theta})d \boldsymbol{\phi}_i d \boldsymbol{\theta} \\
\leq &8 \sigma^2 \log (3 B K)(2 B K)^{2 L+2}\\
&\left\{\left(e_0+\frac{BK}{B K-1}\right)^2+\frac{1}{(2 B K)^2-1}+\frac{2}{(2 B K-1)^2}\right\}\\
\leq &r^n.
\end{aligned}
\end{equation}

Then,
\begin{equation}
\begin{aligned}
\ell_n(P_i,P_{\boldsymbol{\phi}_i, \boldsymbol{\theta}}) =&\frac{1}{2 \sigma_\epsilon^2}\left(\left\|\boldsymbol{y}-f_{\boldsymbol{\phi}_i, \boldsymbol{\theta}}(\boldsymbol{x})\right\|_2^2-\left\|\boldsymbol{y}-f_i(\boldsymbol{x})\right\|_2^2\right) \\
=&\frac{1}{2 \sigma_\epsilon^2}\left(\left\|f_{\boldsymbol{\phi}_i, \boldsymbol{\theta}}(\boldsymbol{x})-f_i(\boldsymbol{x})\right\|_2^2\right.\\
&\left.+2\left\langle \boldsymbol{y}-f_i(\boldsymbol{x}), f_i(\boldsymbol{x})-f_{\boldsymbol{\phi}_i, \boldsymbol{\theta}}(\boldsymbol{x})\right\rangle\right).
\end{aligned}
\end{equation}

Therefore, we have
\begin{equation}
\label{eq.38}
\begin{aligned}
\int_{\Theta}\int_{\Theta} \ell_n(P_i,P_{\boldsymbol{\phi}_i, \boldsymbol{\theta}})\hat{q}(\boldsymbol{\phi}_i,\boldsymbol{\theta})d \boldsymbol{\phi}_i d \boldsymbol{\theta}&=\mathcal{H}_1 / 2 \sigma_\epsilon^2+\mathcal{H}_2 / \sigma_\epsilon^2 \\
& \leq nC' r^n +nC'\sum_{i=1}^N \xi_i^n,
\end{aligned}
\end{equation}
where
\begin{equation*}
\begin{aligned}
& \mathcal{H}_1=\int_{\Theta}\int_{\Theta} \left\|f_{\boldsymbol{\phi}_i, \boldsymbol{\theta}}(\boldsymbol{x})-f_i(\boldsymbol{x})\right\|_2^2\hat{q}(\boldsymbol{\phi}_i,\boldsymbol{\theta})d \boldsymbol{\phi}_i d \boldsymbol{\theta}, \\
& \mathcal{H}_2=\int_{\Theta}\int_{\Theta} \left\langle \boldsymbol{y}-f_i(\boldsymbol{x}), f_i(\boldsymbol{x})-f_{\boldsymbol{\phi}_i, \boldsymbol{\theta}}(\boldsymbol{x})\right\rangle\hat{q}(\boldsymbol{\phi}_i,\boldsymbol{\theta})d \boldsymbol{\phi}_i d \boldsymbol{\theta}.
\end{aligned}
\end{equation*}

As we can see \eqref{eq.35} and \eqref{eq.38} stand for two key terms for the Lemma 1, then combining them, w.h.p.,
\begin{equation*}
\begin{aligned}
\frac{1}{N}\sum_{i=1}^{N}\inf _{q(\boldsymbol{\phi}_i,\boldsymbol{\theta})}\left\{\int_{\Theta}\int_{\Theta} \ell_n(P_i,P_{\boldsymbol{\phi}_i, \boldsymbol{\theta}})q(\boldsymbol{\phi}_i,\boldsymbol{\theta})d \boldsymbol{\phi}_i d \boldsymbol{\theta}\right.\\
\left.+\mathrm{KL}[q(\boldsymbol{\phi}_i, \boldsymbol{\theta}) \| \pi^*(\boldsymbol{\phi}_i, \boldsymbol{\theta})]\right\} \leq nC r^n +\frac{nC'}{N}\sum_{i=1}^N \xi_i^n.
\end{aligned}
\end{equation*}

\section*{Proof of the Lemma 2}

As a important tool for the proof, we first restate the KL divergence from Donsker and Varadhan's representation \cite{boucheron2013concentration} as follows.

\begin{lemma}
\textit{For any probability measure µ and any measurable function $h$ with $e^h \in L_1(\mu)$},
\begin{equation}
\label{eq.40}
\log \int e^{h(\eta)} \mu(d \eta)=\sup _\rho\left[\int h(\eta) \rho(d \eta)-\operatorname{KL}(\rho \| \mu)\right].
\end{equation}

Next, by the Theorem 3.1 of \cite{pati2018statistical}, it shows
\begin{equation}
\label{eq.41}
\int_{\Theta} \int_{\Theta} \eta\left(P_{\boldsymbol{\phi}_i, \boldsymbol{\theta}},P_i\right) \pi^*(\boldsymbol{\phi}_i, \boldsymbol{\theta}) d \boldsymbol{\phi}_i d \boldsymbol{\theta} \leq e^{C'' n (\varepsilon^n)^2} \text {, w.h.p., }
\end{equation}
where 
\begin{equation*}
\log\eta\left(P_{\boldsymbol{\phi}_i, \boldsymbol{\theta}},P_i\right)=\ell_n(P_{\boldsymbol{\phi}_i, \boldsymbol{\theta}},P_i)+nd^2(P_{\boldsymbol{\phi}_i, \boldsymbol{\theta}},P_i).
\end{equation*}
\end{lemma}

Furthermore, combining the Lemma 3 and \eqref{eq.41}, then let $h(\eta)=\log \eta\left(P_{\boldsymbol{\phi}_i, \boldsymbol{\theta}},P_i\right)$, $\mu=\pi^*(\boldsymbol{\phi}_i, \boldsymbol{\theta})$, and $\rho=\hat{q}(\boldsymbol{\phi}_i,\boldsymbol{\theta})$, thus we have
\begin{equation}
\begin{aligned}
\int_{\Theta}\int_{\Theta} &d^2(P_i,P_{\boldsymbol{\phi}_i, \boldsymbol{\theta}})\hat{q}(\boldsymbol{\phi}_i,\boldsymbol{\theta})d \boldsymbol{\phi}_i d \boldsymbol{\theta}\\
\leq &\frac{1}{n} \left\{\int_{\Theta}\int_{\Theta} \ell_n(P_i,P_{\boldsymbol{\phi}_i, \boldsymbol{\theta}})\hat{q}(\boldsymbol{\phi}_i,\boldsymbol{\theta})d \boldsymbol{\phi}_i d \boldsymbol{\theta} \right. \\
&+\mathrm{KL}[\hat{q}(\boldsymbol{\phi}_i, \boldsymbol{\theta}) \| \pi^*(\boldsymbol{\phi}_i, \boldsymbol{\theta})] \bigg\}+C'' (\varepsilon^n)^2.
\end{aligned}
\end{equation}

Therefore, for the all $N$ clients, we have
\begin{equation*}
\begin{aligned}
\frac{1}{N}\sum_{i=1}^{N}\int_{\Theta}\int_{\Theta} &d^2(P_i,P_{\boldsymbol{\phi}_i, \boldsymbol{\theta}})\hat{q}(\boldsymbol{\phi}_i,\boldsymbol{\theta})d \boldsymbol{\phi}_i d \boldsymbol{\theta} \\
\leq &\frac{1}{nN}\sum_{i=1}^{N}\left\{\int_{\Theta}\int_{\Theta} \ell_n(P_i,P_{\boldsymbol{\phi}_i, \boldsymbol{\theta}})\hat{q}(\boldsymbol{\phi}_i,\boldsymbol{\theta})d \boldsymbol{\phi}_i d \boldsymbol{\theta}\right.\\
&+\mathrm{KL}[\hat{q}(\boldsymbol{\phi}_i, \boldsymbol{\theta}) \| \pi^*(\boldsymbol{\phi}_i, \boldsymbol{\theta})]\bigg\}+C'' (\varepsilon^n)^2.
\end{aligned}
\end{equation*}

%Use $\backslash${\tt{appendix}} if you have a single appendix:
%Do not use $\backslash${\tt{section}} anymore after $\backslash${\tt{appendix}}, only $\backslash${\tt{section*}}.
%If you have multiple appendixes use $\backslash${\tt{appendices}} then use $\backslash${\tt{section}} to start each appendix.
%You must declare a $\backslash${\tt{section}} before using any $\backslash${\tt{subsection}} or using $\backslash${\tt{label}} ($\backslash${\tt{appendices}} by itself
 %sstarts a section numbered zero.)}


\bibliography{TNNLS}
\bibliographystyle{ieeetr}

\begin{IEEEbiography}[{\includegraphics[width=1in,height=1.25in,clip,keepaspectratio]{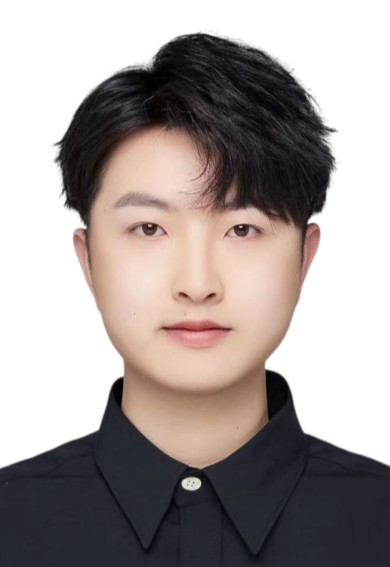}}]{Hui Chen}
is currently a Ph.D. candidate in the School of Computing, Macquarie University. His current research interests include machine learning, Bayesian deep learning, and federated learning. 
\end{IEEEbiography}

\begin{IEEEbiography}[{\includegraphics[width=1in,height=1.25in,clip,keepaspectratio]{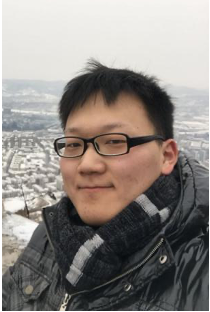}}]{Hengyu Liu} received the B.S. degree in computer science from Northeastern University, China, in 2017. He is currently a Ph.D with computer software and theory of Northeastern University, China. His research interests include Deep Learning, Machine Learning and AI for Education.
\end{IEEEbiography}

\begin{IEEEbiography}[{\includegraphics[width=1in,height=1.25in,clip,keepaspectratio]{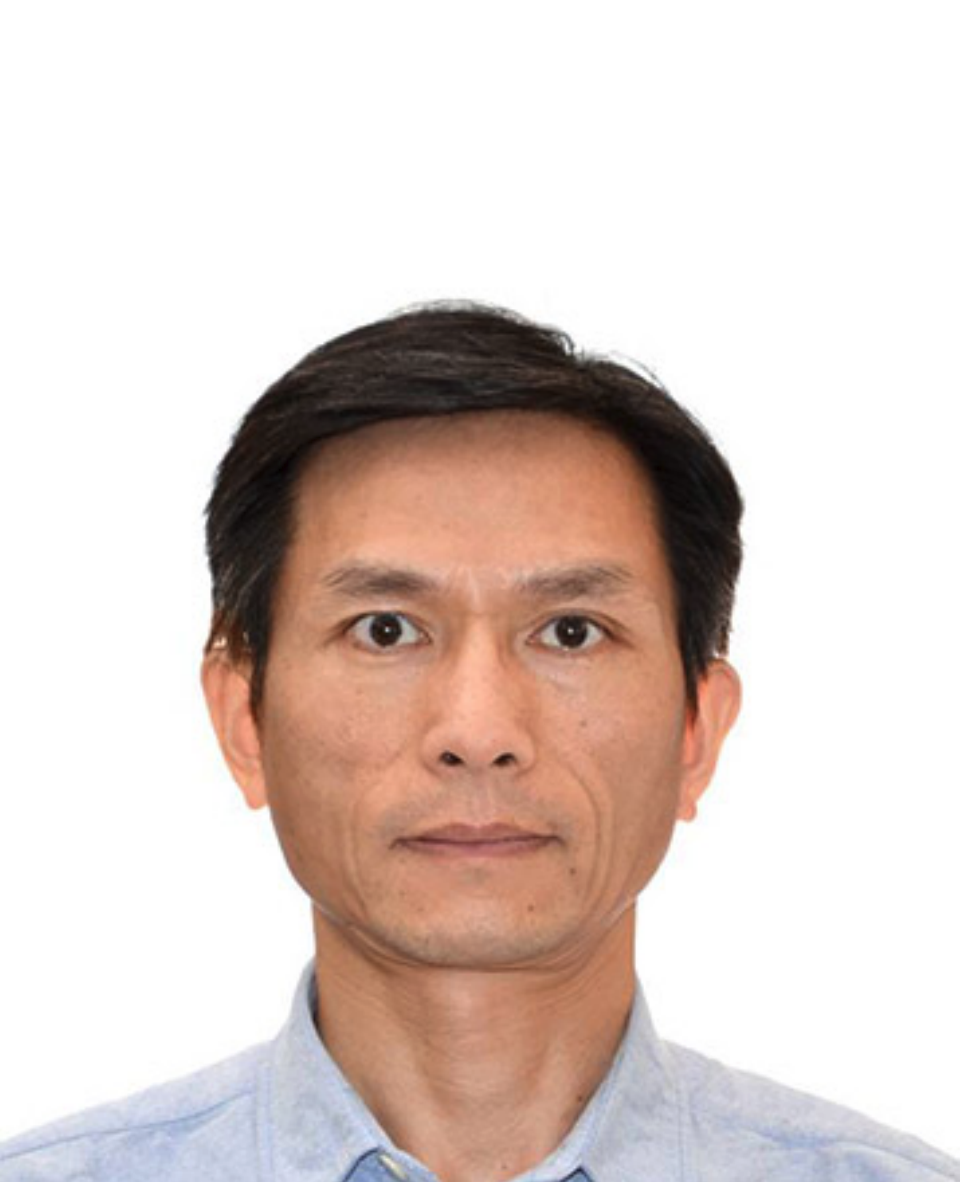}}]{Longbing Cao}
(Senior Member, IEEE) received the
Ph.D. degrees in pattern recognition and intelligent systems and in computing science from the Chinese Academy of Sciences, Beijing, China, in 2002, and the University of Technology Sydney (UTS), Sydney, NSW, Australia, in 2004, respectively.

He is currently a Distinguished Chair in AI at Macquarie University and an ARC Future Fellow (Level 3). His research interests include artificial intelligence, data science, knowledge discovery, machine learning, behavior informatics, complex
intelligent systems, and enterprise innovation. 
\end{IEEEbiography}

\begin{IEEEbiography}[{\includegraphics[width=1in,height=1.25in,clip,keepaspectratio]{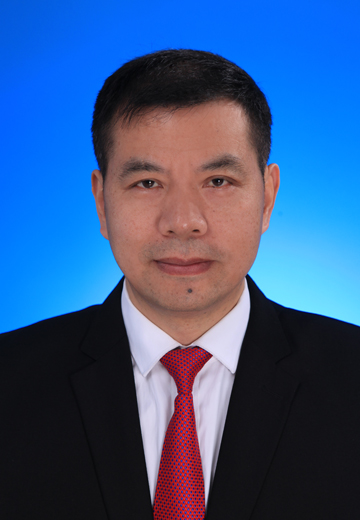}}]{Tiancheng Zhang} received the Ph.D degree in computer software and theory from Northeastern University (NEU) of China. He is currently an associate professor in the School of Computer Science and Engineering of NEU. His research interests include big data analysis, spatio-temporal data management, and deep learning.
\end{IEEEbiography}

 

% \end{IEEEbiography}



\vfill

\end{document}